\documentclass{article}

\usepackage[nonatbib,preprint]{neurips_2024}
\usepackage[square,sort,comma,numbers]{natbib}

\usepackage[utf8]{inputenc} 
\usepackage[T1]{fontenc}    
\usepackage{hyperref}       
\usepackage{url}            
\usepackage{booktabs}       
\usepackage{amsfonts}       
\usepackage{nicefrac}       
\usepackage{microtype}      
\usepackage{xcolor}         
\usepackage{lipsum}
\usepackage{subcaption}
\usepackage{algorithm}
\usepackage{algpseudocode}

\usepackage{graphicx}
\usepackage{multirow}
\usepackage{eccvabbrv}
\usepackage{amsmath} 
\usepackage[permil]{overpic}
\usepackage{iac_pkg}
\usepackage[capitalize]{cleveref}

\title{Perturb, Attend, Detect and Localize (PADL): \\Robust Proactive Image Defense}

\author{%
  Filippo Bartolucci\\
  Computer Science and Engineering Dept.\\
  CVLab, University of Bologna\\
  Bologna, Italy \\
  \texttt{filippo.bartolucci3@unibo.it} \\
  \And
  Iacopo Masi\\
  Computer Science Dept.\\
  OmnAI Lab, Sapienza, University of Rome\\
  Rome, Italy \\
  \texttt{masi@di.uniroma1.it} \\
  \And
  Giuseppe Lisanti\\
  Computer Science  and Engineering Dept.\\
  CVLab, University of Bologna\\
  Bologna, Italy \\
  \texttt{giuseppe.lisanti@unibo.it} \\
}

\begin{document}

\maketitle

\begin{abstract}
Image manipulation detection and localization have received considerable attention from the research community given the blooming of Generative Models (GMs).
Detection methods that follow a \emph{passive} approach may overfit to specific GMs, limiting their application in real-world scenarios, due to the growing diversity of generative models.
Recently, approaches based on a \emph{proactive} framework have shown the possibility of dealing with this limitation. 
However, these methods suffer from two main limitations, which raises concerns about potential vulnerabilities: i) 
the manipulation detector is not robust to noise and hence can be easily fooled; ii) the fact that they rely on fixed perturbations for image protection offers a predictable exploit for malicious attackers, enabling them to reverse-engineer and evade detection.
To overcome this issue we propose PADL, a new solution able to generate image-specific perturbations using a symmetric scheme of encoding and decoding based on cross-attention, which drastically reduces the possibility of reverse engineering, even when evaluated with adaptive attacks~\cite{tramer2020adaptive}.
Additionally, PADL is able to pinpoint manipulated areas, facilitating the identification of specific regions that have undergone alterations, and has more generalization power than prior art on held-out generative models. 
Indeed, although being trained only on an attribute manipulation GAN model~\cite{liu2019stgan}, our method generalizes to a range of unseen models with diverse architectural designs, such as StarGANv2, BlendGAN, DiffAE, StableDiffusion and StableDiffusionXL.
Additionally, we introduce a novel evaluation protocol, which offers a fair evaluation of localisation performance in function of detection accuracy and better captures real-world scenarios.
\end{abstract}

\section{Introduction}\label{sec:intro}
Advancements in Generative Models (GMs) for image synthesis have continually transformed the landscape of the field, showcasing remarkable capabilities in tasks ranging from unconditional image generation from random noise to nuanced manipulation given a natural image to edit. Nevertheless, this progress introduces significant security concerns because a ill-intentioned user could alter the semantics of a genuine image to attain a malicious objective. To address this issue, several counter tools have been developed focusing on binary detection of manipulations~\citep{rössler2019faceforensics, li2019exposing, nirkin2020deepfake, chai2020makes, dang2020detection, zhou2018learning, yang2020constrained} limited to specific GMs. 
Trained on both authentic and manipulated images, these methods are categorized as passive because detection countermeasures are performed \emph{after} manipulation. 
However, their performance and ability to generalize are limited because they need to be retrained for each new GM released, a time-consuming and demanding task given the large number of GMs released every day. 
Recent solutions have addressed the limitations of passive methods by adopting a proactive schema~\citep{asnani2022proactive,asnani2023malp,wang2021faketagger,ruiz2020disrupting} which implements countermeasures \emph{before} any manipulation occurs. 
This new proactive technology intercepts a painful need of the community towards having the right to discern what is generated from what is authentic. 
Indeed, in the private sector, big tech companies of the caliber of Meta and Google dedicated resources to the design of solutions that proactively detect AI-generated contents~\citep{meta_aidetection_2024,google_synthid}.
Various proactive approaches have been proposed so far, among these solutions image tagging~\citep{wang2021faketagger} introduces a hidden message into the image in order to verify the provenance of the image while the solution proposed in~\citep{ruiz2020disrupting} aims directly at disrupting the output of the generative models that are used to manipulate the image. 
Recently, proactive detection techniques~\citep{asnani2022proactive,asnani2023malp} were introduced by augmenting the input image with an additive perturbation\footnote{The additive perturbation is called ``template'' using~\citep{asnani2022proactive,asnani2023malp}'s terminology but we use perturbation in the rest of the paper for clarity.} as a form of protection.
When a protected image is altered, the embedded perturbation is also tampered with, preventing its verification and enabling the detection of manipulations.

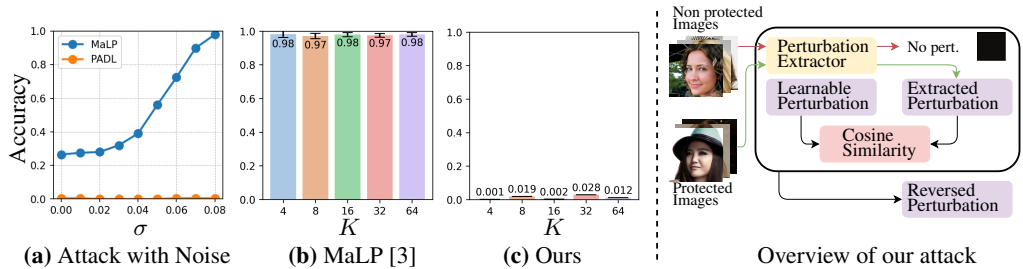
\begin{figure}[t]
    \centering
    \begin{overpic}[width=0.95\textwidth]{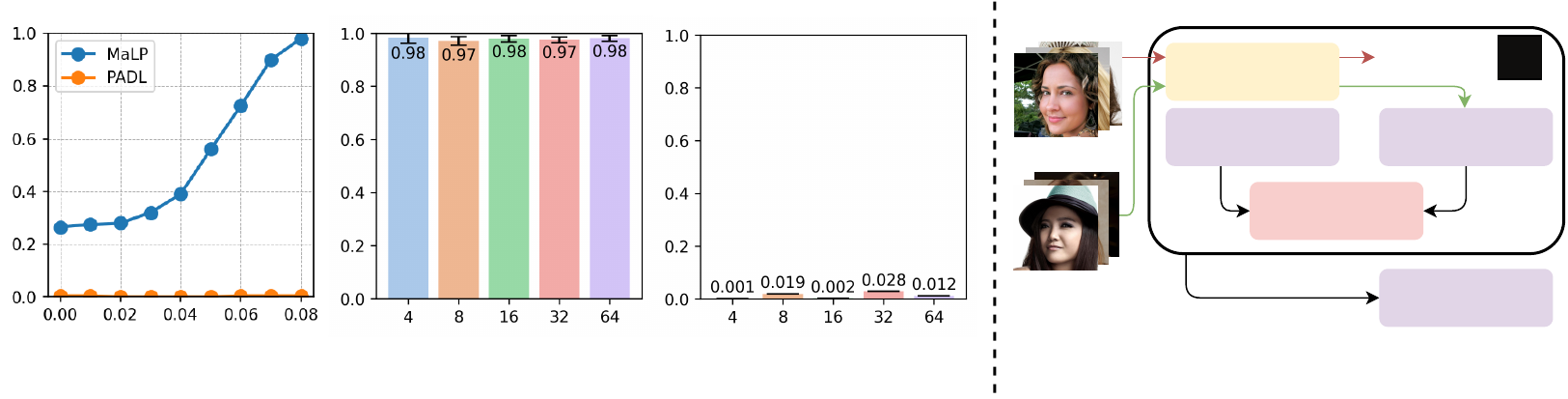}
    
    \put(650,247){\tiny{Non protected}}
    \put(650,235){\tiny{Images}}
    
    \put(650,70){\tiny{Protected}}
    \put(650,58){\tiny{Images}}

    \put(885,212){\tiny{No pert.}}
    \put(755,212){\scriptsize{Perturbation}}
    \put(755,197){\scriptsize{Extractor}}

    \put(885,170){\scriptsize{Extracted}}
    \put(885,155){\scriptsize{Perturbation}}

    \put(820,123){\scriptsize{Cosine}}
    \put(820,108){\scriptsize{Similarity}}

    \put(755,170){\scriptsize{Learnable}}
    \put(755,155){\scriptsize{Perturbation}}

    \put(885,67){\scriptsize{Reversed}}
    \put(885,52){\scriptsize{Perturbation}}
    
    \put(265,0){\small{\textbf{(b)} MaLP~\cite{asnani2023malp}}}%
    \put(480,0){\small{\textbf{(c)} Ours}}%
    \put(0,0){\small{\textbf{(a)} Attack with Noise}}%
    \put(315,28){\small{$K$}}%
    \put(520,28){\small{$K$}}%
    \put(110,28){\small{$\sigma$}}%
    \put(735,0){\small{Overview of our attack}}%
    \put(-15,100){\rotatebox{90}{\small{{Accuracy}}}}
    \end{overpic}  
    \caption{\tbf{Attack to proactive defense methods.}  \emph{left:} 
    \tbf{(a)} When simple Gaussian noise with increasing $\sigma$ is added to an image, the solution from~\cite{asnani2023malp} will detect these images as protected, whereas PADL demonstrates its robustness to noise.
    \tbf{(b)} by using the process on the \emph{right} we were able to reverse engineer the perturbation of~\citep{asnani2023malp} and use it to protect new arbitrary images, achieving a high detection accuracy. 
    \tbf{(c)} we apply the same attack to our solution which remains robust even when a larger collection of protected images is provided.  
    \emph{right:} Our attack uses a fixed set of $K$ protected images to reverse the protection of a proactive method.
    All results have been obtained by averaging over 10 trials.}
    \label{fig:reverse_result}
\end{figure}

However, it is worth highlighting that both~\citep{asnani2022proactive,asnani2023malp} suffer from two main limitations that could potentially be exploited by an attacker. On the one hand, the manipulation detector is not robust to noise and can be easily fooled by simply adding Gaussian noise to an image. However, in this case, tuning the value of $\sigma$  for this noise is not easy as a low value may not fool the detector while a high value may corrupt the image too much. In addition, an attacker usually does not have access to the manipulation detector.
On the other hand, both methods use a fixed set of perturbations and, by reverse engineering one of the predetermined perturbations, an attacker could manipulate images and authenticate them as real using the reversed perturbation.  
To this end, we conducted two experiments, one adding Gaussian noise with increasing $\sigma$ and the second to reverse engineer the perturbation used in~\citep{asnani2023malp}.
Fig.~\ref{fig:reverse_result}(a)(b) shows that this family of solutions can be easily bypassed.

To address this issue, our research aims to enhance the proactive protection mechanism by transitioning from a finite set of perturbations to a customized perturbation per image, which ensures robustness as shown in Fig.~\ref{fig:reverse_result}(a)(c).
However, designing image-specific perturbations is a challenging task due to the lack of a clear ground truth to guide the model. 
To overcome this limitation, we leverage the transformer architecture's cross-attention mechanism, and condition a learnable perturbation on the image, resulting in a unique protection, tailored to the specific characteristics of each input image.
The framework consists of an Encoding and Decoding module with a symmetric cross-attention mechanism. The encoder customizes a sequence of learnable tokens through cross-attention layers to create a personalized perturbation for each image, while the decoder recovers this perturbation to detect and localize manipulations. Additionally, a revamped loss function enforces diversity in the perturbations, contributing to the effectiveness of our approach. With this, protected images can be shared online, and their authenticity can always be verified through the decoder module.

In addition, the performance of state-of-the-art for proactive methods reported~\cite{asnani2023malp} is biased toward manipulated pixels. Indeed, the protocol defined for detection uses both non-manipulated and manipulated images, while the localisation considers only manipulated images. 
This may seem straightforward if detection works perfectly, but as shown in~\cref{tab:generalization}, detection often fails with unseen GMs, making it unreliable to decide when to compute localisation. 
For this reason, we introduce a new evaluation protocol where localisation depends on the detection accuracy. 
This new protocol provides a fair comparison by considering both non-manipulated and manipulated images when evaluating the localisation and better generalizes to real-world scenarios.

The contributions of our work can be summarized as follows:

$\diamond$ 
we empirically demonstrate the vulnerability of state-of-the-art proactive schemes to either noise or to our black-box attack which allows estimating the perturbation from the protected image and adding it to malicious manipulated images so that the proactive scheme will detect them as protected and real.

$\diamond$ 
we introduce PADL a new proactive solution which is robust to our black-box attack and to adaptive attacks~\cite{tramer2020adaptive} specifically tailored for our method.

$\diamond$ 
PADL achieves remarkable generalization capabilities, indeed, although being trained only on STGAN it generalizes to StarGANv2, BlendGAN, DiffAE, StableDiffusion and StableDiffusion XL. 

$\diamond$ 
We define a new evaluation protocol for image manipulation detection and localization that ensures a balanced and realistic comparison. This protocol uses both manipulated and non-manipulated images for localization, and conditions the evaluation on the detector's prediction, thereby reflecting real-world scenarios where most images are authentic.

\section{Related work}\label{sec:related_works}
\minisection{Passive defense} Prior works proposed methods against image manipulation that follow a passive protocol, which means that countermeasures are taken after the manipulation has occurred. 
In this category, earlier methods~\citep{rössler2019faceforensics, li2019exposing} identify artifacts left by generative models in the RGB representation of the image. This was achieved by training models on a dataset of real and manipulated images to discriminate between them by examining the visual content.
Nirkin~\etal~\citep{nirkin2020deepfake} improved manipulation detection methods by exploiting face-context discrepancies. 
The approach integrated a face identification network for precise semantic segmentation and a context recognition network that considered hair, ears and neck.
By utilizing signals from both networks to identify discrepancies, they enhanced traditional fake image detection.
Chai~\etal~\citep{chai2020makes} introduced a patch-based CNN classifier to identify and visualize the regions of an image that have undergone manipulation. The classifier slides through the different image patches to determine if it is real or not, thus verifying if and in which region manipulation has occurred.
Dang~\etal~\citep{dang2020detection} proposed an alternative approach by incorporating an attention mechanism to process and enhance feature maps for the detection task. 
The feature map is then used to highlight informative regions, improving binary classification and visualizing manipulated regions.
New solutions~\citep{zhou2018learning, yang2020constrained} shifted the focus from the image content to the noise present in some regions of the image. Zhou~\etal~\citep{zhou2018learning} utilises RGB images and noise features extracted using steganalysis rich filter model, in conjunction with a Faster R-CNN module, to detect forgeries. 
Similarly, Yang~\etal~\citep{yang2020constrained} followed the same approach yet employed a trainable noise extractor based on Constrained CNN~\citep{pathak2015constrained}. 
This choice was motivated by the susceptibility of previous filters to adversarial attacks.
HiFiNet was proposed in~\citep{guo2023hierarchical} leveraging four branch encoders that learn a fine-grained hierarchical categorization of the manipulation and provide 2D localization for the manipulation.
While the works described above have produced interesting results, these models perform poorly when applied to new manipulations not seen during training: manipulation generated by different techniques can have different visual artifacts, which hampers the generalization of all learning-based passive methods.

\minisection{Proactive defense} To overcome the limitations of passive methods, researchers have started to explore proactive approaches, in the sense that countermeasures are implemented before any manipulation occurs. 
The solution from Ruiz \etal~\citep{ruiz2020disrupting} proposed to disrupt the generator output by applying an imperceptible perturbation to real images. 
The perturbation is generated by a modified version of adversarial attacks such as FGSM, I-FGSM, and PGD and is able to generalize across different image conditioning classes. 
However, this solution does not work in a black-box scenario as it requires knowledge of a specific GM.
Wang~\etal~\citep{wang2021faketagger} introduced a solution closer to the concept of watermarking. 
This approach embeds a hidden message within real images, ensuring its retrieval even after manipulations in order to authenticate the image's identity. 
A U-Net model embeds a bit sequence into the images, leveraging redundancy to enhance resistance against manipulation. 
Although not intended as a detection tool, this technique can be used to track the origin of changes within a social network by linking each user image to its unique identifier. The concept of watermarking has been extended in~\citep{asnani2024promark} by applying proactive watermarking to the train data and then training or fine-tuning a GM to maintain the watermark. This approach enables the extraction of the watermark from newly generated images yet assumes being the ``owner'' of the GM.
Asnani~\etal~\citep{asnani2022proactive} proposed a proactive framework for generalized manipulation detection in which a perturbation is added to the input image. 
If manipulations occur, the perturbation is tampered and the image can be detected as manipulated. 
The perturbation is randomly selected from a finite set that have been learned at training time. 
This work has been successively extended in~\citep{asnani2023malp} with the introduction of manipulation localization. 

In this paper, we show that~\citep{asnani2023malp,asnani2022proactive} are prone to attack, while our method generalizes across diverse unseen GMs and offers a per-image protection perturbation that minimizes the vulnerabilities of predictability caused by the reuse of the same set of perturbations.

\section{Method}
\label{sec:method}
The proposed approach relies on a set of learnable tokens that, conditioned on the input image $\bx$, produces a image-specific perturbation $\pert_e$.
This perturbation is used for the detection and localization of manipulations, employing two primary components: an Encoding Module and a Decoding Module. The encoding part is composed by a perturbation encoder $\pertenc$ that transforms the learnable tokens into a perturbation conditioned on the input image $\bx$.
The decoding part is composed by a protection decoder $\pertdec$ that extracts the perturbation $\pert_d$ and a Map Block $\mathcal{M}$ in charge of performing manipulation detection and localizing the manipulations. 
 
\begin{figure}[tb]
    \centering
        \begin{overpic}[width=.99\textwidth]{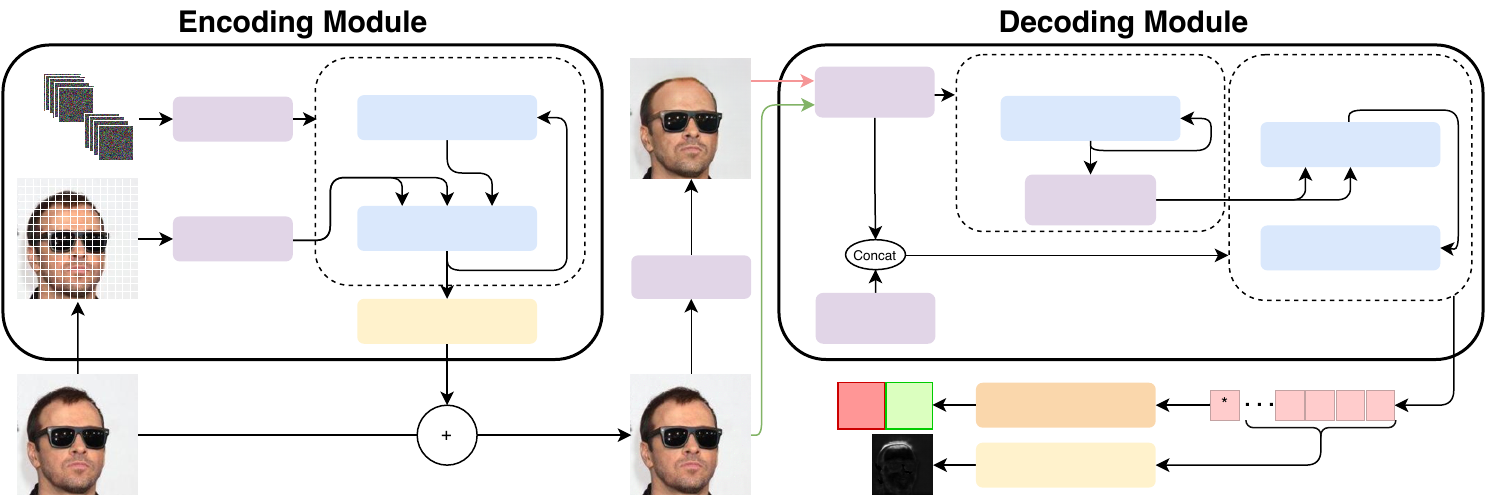}
        \put(22,292){\tiny{Set of tokens}}
        \put(7,-10){\tiny{Real Image $\bx$}}
        \put(255,115){\tiny{Proj + Reshape}}
        \put(390,-10){\tiny{Protected Image $\prot(\bx)$}}
        \put(421,320){\tiny{Manipulated}}
        \put(416,305){\tiny{Image $\mathcal{G}\big(\prot(\bx)\big)$}}

        \put(360,220){\tiny{xN}}
        \put(780,220){\tiny{xN}}
        \put(960,270){\tiny{xN}}
        
        \put(426,147){\tiny{Generator $\gen$}}
        \put(590,-10){\tiny{Map}}
        \put(222,283){\tiny{\textit{Perturbation Encoder $\pertenc$}}}
        \put(242,178){\scriptsize{Cross-Attention}}
        \put(248,254){\scriptsize{Self-Attention}}

        \put(125,259){\tiny{Token}}
        \put(125,244){\tiny{Embedder}}
        
        \put(125,179){\tiny{Patch}}
        \put(125,164){\tiny{Embedder}}

        \put(652,283){\tiny{\textit{Perturbation Decoder $\pertdec$}}}        
        \put(680,252){\scriptsize{Self-Attention}}
        \put(700,205){\tiny{Recovered}}
        \put(696,190){\tiny{Perturbation}}
        \put(557,278){\tiny{Patch}}
        \put(557,263){\tiny{Embedder}}

        \put(557,125){\tiny{Class}}
        \put(557,110){\tiny{Token}}

        \put(564,82){\tiny{Detection}}
        \put(860,283){\tiny{\textit{Map Block $\mathcal{M}$}}}
        \put(849,235){\scriptsize{Cross-Attention}}
        \put(853,165){\scriptsize{Self-Attention}}

        \put(805,78){\tiny{Class Token + Map Token}}
        \put(690,58){\scriptsize{MLP}}
        \put(665,18){\tiny{Proj + Reshape}}
        
    \end{overpic}
    
    \caption{\tbf{Architecture overview.} The encoding module creates a specific perturbation and adds it to a real image for protection. The decoding module first estimates the perturbation and then uses it to perform manipulation detection and localization.} 
    \label{fig:overview}
\end{figure}
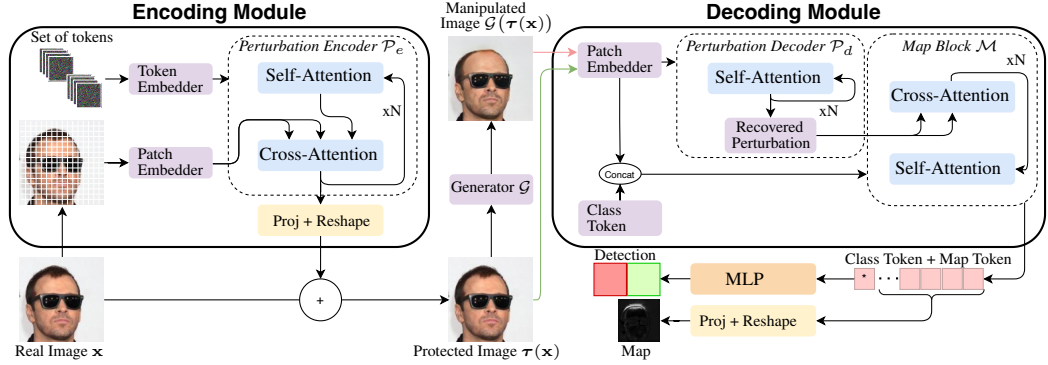

All the components of our architecture, i.e., $\pertenc$, $\pertdec$ and $\mathcal{M}$, consists of $N$ ViT-like transformer blocks.
The whole architecture is shown in \cref{fig:overview}.

\subsection{Encoding image-specific perturbations} \label{sec:encoding_module}
The encoding module is used to protect real images $\bx \in \mathbb{R}^{H\times W\times3}$ before manipulation occurs. 
In passive approaches, detection involves discerning between an authentic image $\bx$ and its manipulated version $\gen(\bx)$, generated by a generative model $\gen$. 
In a proactive framework~\citep{asnani2022proactive,asnani2023malp}, given an authentic image $\bx$, the method applies a transformation $\prot(\cdot)$ to create a protected image $\prot(\bx)$, 
then the manipulation detection process occurs between protected image $\prot(\bx)$ and its manipulated counterpart $\gen\big(\prot(\bx)\big)$. It is important to note that, since the detection process is based solely on the verification of the presence of the perturbation, an image without a perturbation cannot be classified as protected, regardless of its authenticity.
In prior proactive approaches~\citep{asnani2022proactive,asnani2023malp}, protection involved randomly selecting a predefined perturbation from a finite set and embedding it into the image. The process for which the perturbation is embedded in the image is additive, i.e., $\prot(\bx) \doteq \bx + \pert_e$, similarly to what is done for adversarial attacks~\cite{madry2018towards} yet using another procedure.

On the contrary, our approach parametrizes the transformation $\prot(\cdot)$ still with an additive perturbation $\pert_e$ yet using a set of $P$ learnable tokens $\{\tbf{t}_i \in \mathbb{R}^{d} | i=1,\ldots, P\}$, where $d$ is the dimension of the inner representation used by our model~\cite{vit}.
The tokens are shared among all the perturbed images. Yet, unlike prior art, we also specialize the perturbation to be image-specific, which means the perturbation is also conditioned on the input data $\bx$, aiming to prevent a black-box attack that may reverse engineer it. We will empirically prove this claim in the experimental section in~\cref{sec:experiments}, but preliminary results can be appreciated in~\cref{fig:reverse_result}.
Our transformation $\prot(\cdot)$ is defined as follows:
\begin{equation}
    \prot(\bx) = \bx + \alpha \cdot \pert_e(\bx;\tokens), \quad \text{with} \quad \alpha > 0
    \label{eq:transformation}
\end{equation}
where $\alpha$ is a fixed non-negative scalar parameter used to control the strength of the protecting perturbation and $\pert_e(\bx;\tokens)$ returns values in $[-1,\ldots,+1]$ using an hyperbolic tangent function. The values are bounded so that the norm of the perturbation added to the image is limited in the $\ell_2$ norm and upper bounded to $\alpha \sqrt{H\times W}$. This avoids the need for extra losses to minimize the  $\ell_2$ norm of the perturbation making the training simpler with less hyper-parameters than~\cite{asnani2023malp}.

The perturbation encoder $\pertenc$ takes as input the learnable tokens $\mathbf{T}_0=\{\tokens\}$ and applies a series of $N$ parameterized transformations based on transformer~\citep{vaswani2023attention} blocks with both self- and cross-attention. The self-attention mechanism only depends on the input tokens while the cross-attention is used to specialize the tokens, conditioning them on the input image $\bx$. 
In particular, we divide the image into $P$ non-overlapping patches of dimension $p\times p$, such that $P= HW/p^2$,
and embed them using a patch embedder into a sequence of tokens $\{\ptokens\}$, where each $\mbf{x}_i \in \mathbb{R}^{d}$ has the same dimensionality of the learnable tokens and $\bX$ indicates the matrix of the patch embeddings, thus $\bX \in\mathbb{R}^{P \PLH d}$. 

At each of the $N$ perturbation encoder blocks, the learnable tokens are conditioned on the input image employing the patch tokens as context in the cross attention as:
\begin{equation}
\mbf{T}_{n} = \on{softmax}\left[s \big(\mbf{T}^{\star}_{n-1}\mbf{W}_q\mbf{W}_k\bX^{\top} \big)\right] \bX \mbf{W}_v + \mbf{T}^{\star}_{n-1}, \quad n=1\ldots N,\label{eq:ca}
\end{equation}
where $\mbf{T}^{\star}_{n-1}\in\mathbb{R}^{P \PLH d}$ is the matrix containing the tokens processed by the self-attention in the same block,
$\mbf{T}_{n}$ is the matrix updated with the conditioning after~\cref{eq:ca},
$\mbf{W}_q,~\mbf{W}_k,~\mbf{W}_v$ are the query, key and value weights matrices, 
and $s$ is a scaling factor, computed as in~\cite{vaswani2023attention,vit}.
 
$\mbf{T}^{\star}_{n-1}$ is used as query into the cross-attention mechanism while the input image patches $\bX$ are used as context, i.e., they serve as keys and values in \cref{eq:ca}. 
This mechanism determines the level of importance that each learnable token in the query $\tokens$ should attribute to the corresponding image token $\ptokens$, enabling the customization of the learned tokens to the visual characteristics of the image. 
In other words, the final perturbation is constructed by taking convex, linear combination of patch embeddings where the combination is learned through the similarity between learnable tokens and patch embeddings. 
In the last block, when $n=N$, the perturbation is finally attained constraining its values with hyperbolic tangent function: 
$\pert_e(\bx;\tokens) = \tanh\big\{\phi_e \big(\mbf{T}_{N}\big)\big\}$,
where $\phi_e$ projects and reshapes the tokens in order to match the dimensions of the image.

\subsection{Decoding image-specific perturbations for manipulation detection and localization} \label{sec:decryption_module}
The decoding module, shown in~\cref{fig:overview}-right, is employed to detect and localize any manipulations that may have occurred. This module is composed by two parts: (i) a perturbation decoder  $\pertdec$, and (ii) a Map Block $\mathcal{M}$ to perform detection and estimate the manipulation map. 

The decoding module can take either \(\prot(\bx)\) or \(\gen\big(\prot(\bx)\big)\) as input. Additionally, since in real-world scenarios unprotected images may also be observed, we include \(\bx\) as an alternative input. This allows training the module to also detect unprotected inputs, in contrast to~\cite{asnani2023malp,asnani2022proactive}.
This input image is transformed into patch embeddings $\bX$ using a different patch embedder than $\pertenc$ and is fed to the perturbation decoder $\pertdec$ with the intent of recovering the protecting perturbation, if present.
Differently from $\pertenc$ and $\mathcal{M}$, the perturbation decoder $\pertdec$ employs only self-attention layers. 
This time it is the patch embeddings to go as input and the output of $\pertdec$ ---i.e. $\bX^{\star}_N$--- is forced to recover the original perturbation $\pert_e$ using a reconstruction loss, i.e. $\bX^{\star}_N \doteq \pert_d \approx \pert_e$---see~\cref{sec:training}.
The recovered perturbation is subsequently exploited by the Map Block $\mathcal{M}$. 

The Map Block follows an architecture similar to the encoding module yet, interestingly, the conditioning is inverted and the patch embeddings are provided as input. 
These patch embeddings are augmented with a learnable class token $\bxcls$\cite{vit} that we concatenate to the classic patch embedding as
$\{\ptokens,\bxcls\}$ where $\bXcls$ indicates the new patch embedding matrix. We seek for an inductive bias where the $\bxcls$ token stores information about being or not being manipulated.
 
Although both blocks receive the sequence of image patches as input, unlike in the encoding part, $\pert_d$ recovered from $\pertdec$ is used as \emph{context} in the cross-attention mechanism of $\mathcal{M}$ to condition the manipulation map estimation. This choice is symmetric in comparison to the encoding module, where image patches were used in the cross-attention context. However, the rationale behind this is that the estimated perturbation $\pert_d$ is intended to serve as a guide for the subsequent localization of manipulation performed on the image token sequence. Given that the output map should retain the original image's details, the signal is employed to highlight the location where the image has been manipulated. The cross-attention block of the decoding part is thus:
\begin{equation}
\mathbf{X}_{\texttt{[CLS]}{,n}} = \on{softmax}\left[s\big(\bXclsn^{\star}\mbf{W}_q\mbf{W}_k\pert_d^{\top} \big)\right] \pert_d\mbf{W}_v   + \bXclsn^{\star} , \quad n=1\ldots N,
\label{eq:ca-dec}
\end{equation}
where $\bXclsn^{\star}$ is the matrix containing the patch embeddings processed by the self-attention in the same block.

It is worth highlighting that the weights matrices $\mbf{W}_q,~\mbf{W}_k,~\mbf{W}_v$, of $\pertenc, \pertdec$ and $\mathcal{M}$ are not shared. 
In the last decoding block, when $n=N$, the final predicted manipulation mask $\mathcal{M}(\bx)$ will be a convex linear combination of the recovered perturbation $\pert_d$, and the way the combination is decided depends on how the recovered perturbation $\pert_d$ ``attends'' to the patch embedding: 

\begin{equation}
\mathcal{M}(\bx)=\mathbf{X}_{\texttt{[CLS]},{n}} \quad \text{as} \quad \big\{\underbrace{\mbf{x}_{1,n}~\ldots~\mbf{x}_{P,n}}_{\text{localization}}~\underbrace{\bx_{\texttt{[CLS]},{n}}}_{\text{detection}}\big\}~~ \text{when}~~~n=N,
\end{equation}
where $\bx_{\texttt{[CLS]},{N}}$ is extracted and fed to a multi-layer perceptron, supervised with binary labels, as detailed in~\cref{sec:training}. The first ${P}$ tokens, instead, are projected and reshaped in order to match the dimension of the ground-truth manipulation maps on which they are supervised. 
Following prior art~\citep{asnani2023malp,fakelocator}, the ground-truth manipulation map is defined as $\mbf{Y} \doteq \frac{1}{2^8-1}\on{gray}( \vert\prot(\bx) - \gen\big(\prot(\bx)\big)\vert)$ where $\on{gray}(\cdot)$ converts an RGB image into grayscale. Each pixel of the manipulation mask takes a continuous value in $[0,\ldots,1]$ indicating how much a pixel has been manipulated.

\subsection{Training}\label{sec:training}
At train time, all the modules, $\pertenc$, $\pertdec$, and $\mathcal{M}$ are jointly optimized on $\bx, \prot(\bx)$ and $\gen\big(\prot(\bx)\big)$. 
For each forward pass, a real image $\bx$ is provided as input to $\pertenc$ which generates the image-specific perturbation $\pert_e$ to obtain the corresponding protected image $\tau(\bx)=\bx+\pert_e$. 
Following prior art~\cite{asnani2022proactive,asnani2023malp}, in order to simulate possible manipulations by generative models, we employ a single GM to manipulate $\prot(\bx)$, resulting in the manipulated protected image $\gen\big(\prot(\bx)\big)$. 
Both $\tau(\bx)$ and $\gen\big(\prot(\bx)\big)$ are then fed to the decoding module, which extracts the perturbation, performs binary detection and estimates the manipulation map. The overall training process is detailed in \cref{alg:training}

\minisection{Loss objectives}
To force the decoded perturbation $\pert_d$ to be similar to the encoded one we apply a reconstruction loss, $\Loss_{\text{rec}}$, while to maximize the similarity between the ground-truth $\mbf{Y}$ and the estimated $\mathcal{M}(\bx)$ manipulations map we use the cosine distance, as in $\Loss_{\text{map}}$:
\begin{equation}
    \Loss_{\text{rec}} = \cosine{\pert_e}{\pert_d},~~~    \Loss_{\text{map}} = \cosine{\mbf{Y}}{\mathcal{M}(\bx)},~~~
\mathcal{L}_{\text{div}} = \sum_{\substack{i,j=1 \\ i\neq j}}^{B} \max \biggl(\cosines{\pert_e[i]}{\pert_e[j]},0 \biggr)
    \label{eq:losses}
\end{equation}
In addition, to ensure variation within the batch for $\pert_e$, we introduced a perturbation diversity loss, $\Loss_{\text{div}}$. This loss is crucial as it constrains $\pertenc$ to generate a unique signal for each image.
This loss computes the cosine similarity between $\pert_e$ for pairs of images within the batch, ensuring varying perturbations across different images. Without this loss, $\pertenc$ would create a single perturbation: plain cosine similarity was not enough, as the model learned only two distinct perturbations with a cosine similarity of -1. Consequently, within a batch, the mean cosine similarity tended to approach zero due to the compensatory effect between same and opposite perturbation comparisons. To address this, negative values were clamped to zero, effectively removing contributions from pairs with negative similarity and forcing the perturbation to be orthogonal. An ablation study on the importance of the $\mathcal{L}_{\text{div}}$ loss component is provided in \cref{sec:ablation}.

Finally, in order to train $\mathcal{M}$ to perform manipulation detection, we simply apply binary cross-entropy to the output of the multi-layer perceptron that processes $\bx_{\texttt{[CLS]},{N}}$ supervised by the binary labels indicating if we are processing $\tau(\bx)$ (protected) vs $\bx$ or $\gen(\prot(\bx))$ (unprotected or manipulated).
In addition, we randomly sum a small Gaussian noise to the image $\bx$ provided as input to the detection loss during training. 
By doing so, we explicitly force our model to distinguish our perturbation from noise applied to the images enabling a more robust protection.

The overall loss employed to optimize the model is given by the sum of all previous terms as
\begin{equation}
\mathcal{L} = \mathcal{L}_{rec} + \mathcal{L}_{map} + \mathcal{L}_{div} + \mathcal{L}_{BCE}
\label{eq:final_loss}
\end{equation}
\begin{algorithm}[!h]
\caption{Training Process}\label{alg:training}
\begin{algorithmic}
\Require $\text{iterations} > 0, \quad \alpha > 0$
\While{$\text{i} < \text{iterations}$}
    \State $\bx \gets \text{dataset.next()}$
    \State $\prot(\bx) = \bx + \alpha \cdot \pert_e(\bx;\tokens)$ \Comment{Encoding}
    \;
    \State $\mathcal{G(}\tau(\bx)) \gets STGAN(\tau(\bx))$ \Comment{GM Manipulation }
    \State $Y \gets |\prot(\bx)-\mathcal{G(}\tau(\bx))| $  \Comment{Ground Truth}

    \;
    \State $\Delta_{\mathcal{G}} = \mathcal{P}_d(\mathcal{G}(\prot(\bx)))$ \Comment{Decoding manipulated images}
    \State $\mathcal{M}, \texttt{[CLS]}_{\mathcal{G}}= \mathcal{M}(\prot(\bx), \Delta_\mathcal{G})$ 

    \;
    \State $\Delta_{\prot(\bx)} = \mathcal{P}_d(\prot(\bx))$ \Comment{Decoding  protected images}
    \State $\_, \texttt{[CLS]}_{\prot(\bx)}= \mathcal{M}(\prot(\bx), \Delta_{\prot(\bx)})$ 

    \;
    \State $\Delta_{\bx} = \mathcal{P}_d(\bx)$  \Comment{Decoding of real images}
    \State $\_, {\texttt{[CLS]}_\bx}= \mathcal{M}(\prot(\bx), \Delta_{\bx})$ 

    \;
    \State $\mathcal{L}= \mathcal{L}_{rec}(\Delta_e, \Delta_{\prot(\bx)}) + \mathcal{L}_{map}(Y, \mathcal{M}) + \mathcal{L}_{div}(\Delta_e) + \mathcal{L}_{BCE}({\texttt{[CLS]}_\bx}, \texttt{[CLS]}_{\prot(\bx)}, \texttt{[CLS]}_{\mathcal{G}}) $
    \State optimizer.step()

\EndWhile
\end{algorithmic}
\end{algorithm}

\subsection{Image protection, manipulation detection and localization}
The proposed approach is consistent with prior works that utilize a proactive method for defending against image manipulation. This method is applicable to any individual or organization, such as journalists and media outlets, that wish to safeguard the integrity of their images. For news agencies publishing sensitive content, such as reports on political events or social unrest, the ability to verify whether an image has been altered is critical in preventing the spread of misinformation. A protected image can be shared online, and its authenticity can always be verified by the decoder module, ensuring that any manipulation or tampering becomes detectable. This process is shown in \cref{alg:inference}.

Additionally, in legal or forensic investigations, where image evidence is crucial, this approach offers an extra layer of security. By embedding an invisible protection, law enforcement agencies and legal professionals can ensure that the images presented in court remain untampered from capture to presentation.

\begin{algorithm}[t]
\caption{Inference Process}\label{alg:inference}
\begin{algorithmic}
\Require $\bx \text{: real image}$
    \State $\prot(\bx) = \bx + \alpha \cdot \pert_e(\bx;\tokens)$ \Comment{Encoding}\\
    Image $\prot(\bx)$ goes ``in the wild''\\ 
    A GM may or not may edit it\\
    Below we assume the manipulation does not occur
    \;
    \State $\Delta_d = \mathcal{P}_d(\prot(\bx))$  \Comment{Decoding}  
    \State $M, \bx_{\texttt{[CLS]}}= \mathcal{M}(\prot(\bx), \Delta_d)$ 
\end{algorithmic}
\end{algorithm}

\section{Experiments}
\label{sec:experiments}

\minisection{Datasets}
Our models have been trained only on the CelebA~\citep{liu2015deep} dataset. 
The images of the dataset have been aligned, centered and cropped to a resolution of $H=W=128$, as in~\citep{asnani2023malp,liu2019stgan}.
During training these images are observed with or without manipulations. 
The manipulated version is generated using only STGAN~\citep{liu2019stgan} which is set to alter the baldness and smile attributes. 
For evaluating the generalization capability of our model to unseen GMs we consider StarGanV2~\cite{choi2020stargan} and four more recent generative models, namely, BlendGAN~\cite{liu2021blendgan} based on generative adversarial networks~\citep{goodfellow2014generative}, DiffAE~\citep{diffae} based on denoising diffusion implicit models~\citep{song2022denoising}, StableDiffusion (SD)~\citep{SD15} and StableDiffusionXL (SDXL)~\citep{sdxl}, both based on latent diffusion models~\citep{rombach2022highresolution}.
In addition, we report in the supplementary material an experiment using a diffusion-based model, i.e., StableDiffusion 1.5, to manipulate the image during training and evaluate the generalization performance also for this model. Using different GMs for training does not impact the generalization performance of PADL, which means the generalization is invariant to the two different GMs used.

As the test set, we employ the subsets of CelebA-HQ and Summer2Winter~\cite{summer2winter} provided in the benchmark defined by~\citep{asnani2023malp}. 
To further extend the evaluation, we selected an additional test set of 200 real images from FFHQ~\citep{karras2019stylebased}. 
The supplementary material provides a list of the GMs used in the evaluation along with the tasks they were used for (e.g., image2image, style transfer, attribute manipulation).

\minisection{Metrics and Evaluation}
To evaluate our model's ability to accurately detect manipulations, we compute the accuracy considering both manipulated and non-manipulated images.
With regard to localization, we compute the Area under the ROC Curve (AUC) between the ground-truth and the estimated manipulation maps. In light of the fact that the ground truth map is still a continuous map, it is necessary to threshold it to calculate the ROC curve. To ensure the absence of any bias in the selection of the threshold, the performance is shown considering different thresholding values, i.e., $t=[0.1, 0.25, 0.5]$.

The performance reported by the state of the art~\cite{asnani2023malp} is biased toward manipulated pixels. More in detail, the protocol defined for detection~\cite{asnani2023malp}, uses 400 images, 200 non-manipulated and 200 manipulated by the GM. While the localisation evaluation uses only the 200 manipulated images. 
This may seem straightforward if detection works perfectly, but as shown in~\cref{tab:generalization}, detection often fails with unseen GMs, making it unreliable to decide when to compute localisation. 
For this reason, we introduce a new evaluation protocol where localisation depends on the detection accuracy. 
This new protocol provides a fair comparison by considering both non-manipulated and manipulated images when evaluating the localisation. This approach balances the two classes and better reflects real-world scenarios, where most images are likely authentic.
In the proposed evaluation protocol, the localization is conditioned on the detector's prediction, and the metrics are calculated for the four following scenarios:
\begin{itemize}
    \item \textbf{Manipulated image correctly detected as manipulated}: Localization evaluation is computed between the ground truth (GT) map and the predicted map, as typically done.
    \item \textbf{Manipulated image incorrectly detected as non-manipulated}: Metrics are computed between the GT map and an all-zero map (indicating "non-manipulation"), reflecting the detection result.
    \item \textbf{Non-manipulated image correctly detected as non-manipulated}: Localization is evaluated between an all-zero GT map (indicating a real image) and a predicted all-zero map.
    \item \textbf{Non-manipulated image incorrectly detected as manipulated}: Metrics are computed between the predicted map and an all-zero GT map (indicating "non-manipulation").
\end{itemize}

With this new evaluation protocol, the localization is conditioned on the accuracy of the detection yet all the methods will be evaluated on the same number of pixels.

\minisection{Implementation details}
The dimension of the patch processed by the transformer has been set to $p=8$. All attention blocks have the same dimensions and number of heads, which were set to 64 and 8, respectively, therefore the dimension of the inner representation used by the transformer is $d=512$.  
The learnable tokens $\tokens$ are initialized with random normal values. 
For both the encoding and the decoding modules we employ learnable positional encodings, as in~\citep{vit}.
For all the experiments the strength of the perturbation $\alpha$ has been set to $0.03$.
We employed the AdamW~\citep{loshchilov2019decoupled} optimizer with an initial learning rate of $1 \times 10^{-4}$.  All models were implemented in PyTorch~\citep{paszke2019pytorch} and trained on an RTX A6000 GPU with 48GB of memory. The total runtime for the training ranges from over $4$ hours for the model with $N=3$ layers to about $10$ hours for the model with $N=12$.

The average time required to protect an image, that is a forward pass through the encoder to generate the perturbation and add it to the image, is $6.93$ ms. 
Conversely, to recover the perturbation and detect if the image has been manipulated, the decoder takes on average $10.27$ ms. Measures are taken on an NVIDIA A6000 synchronizing all cuda events.

\begin{table}[ht]
    \centering
    \caption{\textbf{Quantitative comparison of manipulated pixels across different GMs}. The table reports the total number of manipulated pixels (absolute), the average number of manipulated pixels per image (mean), and the percentage of manipulated pixels relative to the total number of pixels.}
    \resizebox{0.85\textwidth}{!}{
    \begin{tabular}{cccc}
        \toprule
        {\textbf{Model}} & 
        {\textbf{Manipulated Pixels}} &
        {\textbf{Average Manipulated Pixels}} & 
        {\textbf{Manipulated Pixels \%}} \\
        \midrule
        StarGANv2 & 1586875 & 7934 & 48\% \\
        BlendGAN  & 1506646 & 7533 & 46\% \\
        DiffAE    & 199591  & 997  &  6\% \\
        SD 1.5    & 328208  & 1641 & 10\% \\
        SDXL      & 426259  & 2131 & 13\% \\
        \bottomrule
    \end{tabular}
    }
\label{tab:man_magnitude}
\end{table}

\subsection{PADL performance across diverse GMs}
Proactive schemes were introduced to generalize the manipulation detection capability of a model to unseen GMs. To this end we evaluated the performance of PADL with different configurations of $N = [3,6,12]$ with GMs and datasets unseen during training.

From \cref{tab:padl_performance} it is evident that the performance of PADL when trained solely on CelebA is generally consistent across various configurations of $N$, particularly when applied to most unseen Generative Models (GMs). For StarGANv2, BlendGAN, SD 1.5, and SDXL, PADL shows high detection accuracy with $N = [3, 6, 12]$, indicating that these GMs manipulations are aggressive enough to be  detected by all models. However, when evaluated on DiffAE, PADL performs less effectively. DiffAE poses the greatest challenge due to its subtle pixel-level manipulations, which result in the lowest generalization performance across all configurations of $N$. DiffAE's lower performance can be attributed to its minimal pixel alterations, which are harder for PADL to detect. This is further corroborated by the results in \cref{tab:man_magnitude}, where we computed the sum of all the pixels considering the soft non-binarized ground-truth masks across GMs. It can be seen that DiffAE is the one that yields the lowest sum by a large margin, proving that it does create subtle manipulations. Interestingly, increasing the value of $N$ to 6 or 12 shows some improvement in detecting these subtle manipulations in DiffAE, likely due to the increased complexity of the perturbations generated by PADL as $N$ grows. As also noted in \cref{fig:pert_depth} in the supplementary material, an increase in $N$ induces a more complex learned perturbation in \cref{eq:transformation}. This gain with DiffAE can be easily explained: as the parameter $N$ induces more complex perturbations  if we stick to small $N$, the perturbation will be coarse and the subtle manipulation of DiffAE will not be strong enough to corrupt the PADL perturbation, thus the PADL decoder will find the manipulated images still “protected” (false negatives). If we increase the perturbation complexity ($N=6$), the PADL decoder is now able to spot the corruption induced by the subtle DiffAE manipulation resulting in a higher detection rate. 

In light of this analysis, for all subsequent experiments, we considered PADL with $N=6$ since in this setting it can detect both subtle manipulations (DiffAE) and other more aggressive GMs for a better coverage of unseen GMs and improved generalization.

\begin{table}[h]
    \centering
    \caption{\tbf{Performance comparison of PADL models with different configurations of $N$.} Each GM has been utilised in combination with all compatible datasets, namely FFHQ and CelebA-HQ (C-HQ).
    ``t'' represents the threshold used to binarize the GT masks. The best results
    are reported in \textbf{bold}, while the second best are \underline{underlined}. For our solution, results with an increasing number of layers $N$ are also provided.}
    \small
    \resizebox*{0.95\textwidth}{!}{
    \begin{tabular}{ccccccc}
        \toprule
        \multirow{2}{*}{\textbf{Model}} & \multirow{2}{*}{\textbf{Dataset}} & \multicolumn{3}{c}{\textbf{Localization}} & \multicolumn{2}{c}{\textbf{Detection}} \\
        \cmidrule(lr){3-5} \cmidrule(lr){6-7}
        &  & AUC t=0.1 ($\uparrow$) & AUC t=0.25 ($\uparrow$) & AUC t=0.5 ($\uparrow$) & Acc. ($\uparrow$) & AUC ($\uparrow$) \\

         \midrule
        
        & &  \multicolumn{5}{c}{\tbf{StarGANv2}} 
        \\
        \cmidrule(lr){1-2}
        \cmidrule(lr){3-5}
        \cmidrule(lr){6-7}
        PADL $N=3$ & C-HQ & \textbf{0.939} & \textbf{0.876} & \textbf{0.848} & \textbf{1.00} & \textbf{1.00} \\
        PADL $N=6$ & C-HQ & 0.933 & 0.868 & 0.835 & 0.985 & \textbf{1.00}\\
        PADL $N=12$ & C-HQ & \underline{0.938} & \underline{0.873} & \underline{0.844} & \underline{0.995} & \underline{0.999}\\

        \midrule
        PADL $N=3$& FFHQ & \textbf{0.951} & \textbf{0.874} & \underline{0.813} & \textbf{0.998} & \textbf{1.00} \\
        PADL $N=6$& FFHQ & 0.933 & \underline{0.868} & \textbf{0.835} & 0.975 & \textbf{1.00}\\
        PADL $N=12$ & FFHQ & \underline{0.943} & \underline{0.868} & 0.808 & \underline{0.985} & \underline{0.999} \\

        \midrule

        &&\multicolumn{5}{c}{\tbf{BlendGAN}} \\
        \cmidrule(lr){1-2}
        \cmidrule(lr){3-5}
        \cmidrule(lr){6-7}
        PADL $N=3$ & C-HQ & \textbf{0.941} &\textbf{ 0.871} & \textbf{0.798} & \textbf{1.00}  & \textbf{1.00} \\
        PADL $N=6$ & C-HQ & \underline{0.937} & 0.864 & 0.789 & \underline{0.997}  & \textbf{1.00}  \\
        PADL $N=12$ & C-HQ & 0.936 & \underline{0.867} & \underline{0.796} & \underline{0.997} & \underline{0.999} \\
        \midrule
        PADL $N=3$ & FFHQ & \textbf{0.943} & \underline{0.854} & \underline{0.792} & \textbf{1.00} & \textbf{1.00} \\
        PADL $N=6$ & FFHQ & \underline{0.940} & \textbf{0.855} & \textbf{0.798} & \underline{0.995} & \textbf{1.00} \\
        PADL $N=12$ & FFHQ& \textbf{0.943} & 0.853 & 0.789 & \textbf{1.00} & \textbf{1.00} \\

         \midrule

        &&\multicolumn{5}{c}{\tbf{DiffAE}} \\
        \cmidrule(lr){1-2}
        \cmidrule(lr){3-5}
        \cmidrule(lr){6-7}
        PADL $N=3$ & C-HQ & 0.704 & 0.688 & 0.651 & \underline{0.882} & \textbf{0.991} \\
        PADL $N=6$ & C-HQ & \textbf{0.757} & \textbf{0.733} & \textbf{0.723} & \textbf{0.908} & 0.984 \\
        PADL $N=12$ & C-HQ & \underline{0.726} & \underline{0.695} & \underline{ 0.692} & 0.835 & \underline{0.983} \\

        \midrule
        PADL $N=3$ & FFHQ & 0.727 & 0.720 & 0.714 & 0.884 & \underline{0.969} \\
        PADL $N=6$ & FFHQ & \textbf{0.762} & \textbf{0.759} & \textbf{0.775} & \textbf{0.926} & 0.965\\
        PADL $N=12$ & FFHQ & \underline{0.750} & \underline{0.741} & \underline{0.731} & \underline{0.913} & \textbf{0.980}\\

         \midrule

        &&\multicolumn{5}{c}{\tbf{SD 1.5}} \\
        \cmidrule(lr){1-2}
        \cmidrule(lr){3-5}
        \cmidrule(lr){6-7}
        PADL $N=3$ & C-HQ & \underline{0.791} & \underline{0.769} & \textbf{0.783} & \textbf{1.00} & \textbf{1.00} \\
        PADL $N=6$ & C-HQ & \textbf{0.794} & 0.766 & \underline{0.775} & \underline{0.997} & \underline{0.999}\\
        PADL $N=12$ & C-HQ & 0.769 & \textbf{0.771} & 0.771  & 0.970 & 0.995 \\

        \midrule
        PADL $N=3$ & FFHQ & \textbf{0.811} & \textbf{0.780} & \underline{0.789} & \textbf{0.998} & \textbf{0.999 }\\
        PADL $N=6$ & FFHQ & \underline{0.808} & 0.774 & 0.779 & 0.980 & \underline{0.994}\\
        PADL $N=12$ & FFHQ & \textbf{0.811} & \underline{0.777} & \textbf{0.781} & \underline{0.990} & 0.989\\

        \midrule

        &&\multicolumn{5}{c}{\tbf{SDXL}} \\
        \cmidrule(lr){1-2}
        \cmidrule(lr){3-5}
        \cmidrule(lr){6-7}
        PADL $N=3$ & C-HQ & \underline{0.810} & \underline{0.770} & \underline{0.774} & \textbf{1.00} & \textbf{1.00} \\
        PADL $N=6$ & C-HQ & \textbf{0.812} & \textbf{0.773} & \textbf{0.776} & \underline{0.997} & \underline{0.999}\\
        PADL $N=12$ & C-HQ & 0.769 & 0.737 & 0.747 & 0.950 & 0.995\\

        \midrule
        PADL $N=3$ & FFHQ & 0.825 & \textbf{0.782} & \textbf{0.782} & \textbf{0.995} & \textbf{0.999} \\
        PADL $N=6$ & FFHQ & \underline{0.827} & 0.776 & 0.774 & 0.970 & \underline{0.995}\\
        PADL $N=12$ & FFHQ & \textbf{0.829} & \underline{0.778} & \underline{0.781} & \underline{0.990} & 0.990\\

        \bottomrule
    \end{tabular}
    
    \label{tab:padl_performance}
}
\end{table}

\subsection{Performance across diverse GMs and comparison with state-of-the-art}
We evaluated the performance of both passive and proactive solutions with GMs and datasets unseen during training.  

\begin{table}[]
    \centering
    \caption{\tbf{Performance comparison with existing solutions across diverse GMs.} Each GM has been utilised in combination with all compatible datasets, namely FFHQ, CelebA-HQ (C-HQ) and Summer2Winter (S2W). The S2W  dataset was employed exclusively with LatentDiffusion models, given that they are the sole model capable of processing non-face images. ``t'' represents the threshold used to binarize the GT masks. The best results
    are reported in \textbf{bold}, while the second best are \underline{underlined}. (*) The solution from~\cite{guo2023hierarchical} employed images manipulated by StarGANv2 during training.}
    \small
    \resizebox*{!}{0.90\textheight}{
    \begin{tabular}{ccccccc}
        \toprule
        \multirow{2}{*}{\textbf{Model}} & \multirow{2}{*}{\textbf{Dataset}} & \multicolumn{3}{c}{\textbf{Localization}} & \multicolumn{2}{c}{\textbf{Detection}} \\
        \cmidrule(lr){3-5} \cmidrule(lr){6-7}
        &  & AUC t=0.1 ($\uparrow$) & AUC t=0.25 ($\uparrow$) & AUC t=0.5 ($\uparrow$) & Acc. ($\uparrow$) & AUC ($\uparrow$) \\

         \midrule
        
        & &  \multicolumn{5}{c}{\tbf{StarGANv2}} 
        \\
        \cmidrule(lr){1-2}
        \cmidrule(lr){3-5}
        \cmidrule(lr){6-7}
        FFD~\citep{dang2020detection} & C-HQ & 0.873 & 0.801 & 0.770 & 0.977 & \underline{0.999}\\
        HiFi~\cite{guo2023hierarchical}(*) & C-HQ & \textbf{0.999} & \textbf{0.999} & \textbf{0.999} & 0.938 & 0.985\\
        MaLP~\cite{asnani2023malp} & C-HQ & 0.883 & 0.775 & 0.663 & \textbf{0.996} & 0.997 \\
        PADL  & C-HQ & 0.933 & 0.868 & 0.835 & 0.985 & \textbf{1.00}\\

        \midrule
        FFD~\citep{dang2020detection} & FFHQ &0.498 & 0.488 & 0.497 & 0.510 & 0.483\\
        HiFi~\cite{guo2023hierarchical}(*) & FFHQ & \textbf{0.999} & \textbf{0.999} & \textbf{0.999} & \textbf{0.997} & \textbf{1.00} \\
        MaLP~\cite{asnani2023malp} & FFHQ & 0.894 & 0.798 & 0.720 & \underline{0.995} & \underline{0.995} \\
        PADL & FFHQ & 0.933 & \underline{0.868} & \underline{0.835} & 0.975 & \textbf{1.00}\\

        \midrule
        &&\multicolumn{5}{c}{\tbf{BlendGAN}} \\
        \cmidrule(lr){1-2}
        \cmidrule(lr){3-5}
        \cmidrule(lr){6-7}
        FFD~\citep{dang2020detection} & C-HQ & \underline{0.857} & \underline{0.778} & \underline{0.755} & \underline{0.975} & \underline{0.994} \\
        HiFi~\cite{guo2023hierarchical} & C-HQ & 0.528 & 0.528 & 0.528 & 0.520 & 0.624\\
        MaLP~\cite{asnani2023malp} & C-HQ & 0.669 & 0.625 & 0.573 & 0.700 & 0.700 \\
        PADL & C-HQ & \textbf{0.937} & \textbf{0.864} & \textbf{0.789} & \textbf{0.997}  & \textbf{1.00}  \\

        \midrule
        FFD~\citep{dang2020detection} & FFHQ & 0.662 & \underline{0.630} & \underline{0.618} & 0.650 & 0.645\\
        HiFi~\cite{guo2023hierarchical} & FFHQ & 0.498 & 0.498 & 0.498 & 0.498 & \underline{0.900} \\
        MaLP~\cite{asnani2023malp} & FFHQ & \underline{0.664} & 0.624 & 0.589 & \underline{0.698} & 0.697 \\
        PADL N=6 & FFHQ & \textbf{0.940} & \textbf{0.855} & \textbf{0.798} & \textbf{0.995} & \textbf{1.00} \\

         \midrule

        &&\multicolumn{5}{c}{\tbf{DiffAE}} \\
        \cmidrule(lr){1-2}
        \cmidrule(lr){3-5}
        \cmidrule(lr){6-7}
        FFD~\citep{dang2020detection} & C-HQ & 0.500 & 0.500 & 0.500 & 0.500 & 0.552 \\
        HiFi~\cite{guo2023hierarchical} & C-HQ & 0.542 & 0.542 & 0.542 & 0.543 & \underline{0.668} \\
        MaLP~\cite{asnani2023malp} & C-HQ & \underline{0.555} & \underline{0.560} & \underline{0.565} & \underline{0.555} & 0.565 \\
        PADL & C-HQ & \textbf{0.757} & \textbf{0.733} & \textbf{0.723} & \textbf{0.908} & \textbf{0.984} \\

        \midrule
        FFD~\citep{dang2020detection} & FFHQ &0.398 & 0.407 & 0.449 & 0.294 & 0.126 \\
        HiFi~\cite{guo2023hierarchical} & FFHQ & \underline{0.582} & \underline{0.581} & \underline{0.581} & \underline{0.588} &  \underline{0.668} \\
        MaLP~\cite{asnani2023malp} & FFHQ & 0.563 & 0.567 & 0.570 & 0.565  & 0.575 \\
        PADL & FFHQ & \textbf{0.762} & \textbf{0.759} & \textbf{0.775} & \textbf{0.926} & \textbf{0.965}\\
         \midrule

        &&\multicolumn{5}{c}{\tbf{SD 1.5}} \\
        \cmidrule(lr){1-2}
        \cmidrule(lr){3-5}
        \cmidrule(lr){6-7}
        FFD~\citep{dang2020detection} & C-HQ & 0.550 & 0.538 & \underline{0.618} & 0.618 & 0.727 \\
        HiFi~\cite{guo2023hierarchical} & C-HQ & 0.503 & 0.503 & 0.503 & 0.502 & \underline{0.794} \\
        MaLP~\cite{asnani2023malp} & C-HQ & \underline{0.620} & \underline{0.592} & 0.563 & \underline{0.667} & 0.668 \\
        PADL & C-HQ & \textbf{0.794} & \textbf{0.766} & \textbf{0.775} & \textbf{0.997} & \textbf{0.999}\\

        \midrule
        FFD~\citep{dang2020detection} & FFHQ & \underline{0.590} & 0.435 & \underline{0.547} & 0.538 & 0.578 \\
        HiFi~\cite{guo2023hierarchical} & FFHQ & 0.499 & 0.499 & 0.499 & 0.500 & \underline{0.755}  \\
        MaLP~\cite{asnani2023malp} & FFHQ & 0.550 & \underline{0.542} & 0.539 & \underline{0.683} & 0.563 \\
        PADL & FFHQ & \textbf{0.808} & \textbf{0.774} & \textbf{0.779} & \textbf{0.980} & \textbf{0.994}\\

        \midrule
        FFD~\citep{dang2020detection} & S2W & 0.371 & 0.400 & 0.414 & 0.333 & 0.077\\
        HiFi~\cite{guo2023hierarchical} & S2W & \underline{0.667} & \underline{0.667} & \underline{0.667} & \underline{0.655} & \underline{0.911} \\
        MaLP~\cite{asnani2023malp} & S2W & 0.613 & 0.583 & 0.566 & 0.637 & 0.638\\
        PADL & S2W & \textbf{0.77} & \textbf{0.741} & \textbf{0.739} & \textbf{0.860} & \textbf{0.910} \\

        \midrule

        &&\multicolumn{5}{c}{\tbf{SDXL}} \\
        \cmidrule(lr){1-2}
        \cmidrule(lr){3-5}
        \cmidrule(lr){6-7}
        FFD~\citep{dang2020detection} & C-HQ & 0.551 & 0.530 & \underline{0.615} & \underline{0.618} & 0.724\\
        HiFi~\cite{guo2023hierarchical} & C-HQ & 0.503 & 0.503 & 0.503 & 0.503 & \underline{0.757}\\
        MaLP~\cite{asnani2023malp} & C-HQ & \underline{0.638} & \underline{0.615} & 0.599 & 0.695 & 0.695 \\
        PADL & C-HQ & \textbf{0.812} & \textbf{0.773} & \textbf{0.776} & \textbf{0.997} & \underline{0.999}\\

        \midrule
        FFD~\citep{dang2020detection} & FFHQ & 0.525 & 0.529 & 0.547 & 0.512 & 0.528 \\
        HiFi~\cite{guo2023hierarchical} & FFHQ & 0.498 & 0.498 & 0.498 & 0.500 & 0.629 \\
        MaLP~\cite{asnani2023malp} & FFHQ & \underline{0.607} & \underline{0.584} & \underline{0.567} & \underline{0.708} & \underline{0.699} \\
        PADL & FFHQ & \textbf{0.827} & \textbf{0.776} & \textbf{0.774} & \textbf{0.970} & \textbf{0.995}\\

        \midrule
        FFD~\citep{dang2020detection} & S2W & 0.371 & 0.401 & 0.414 & 0.333 & 0.114\\
        HiFi~\cite{guo2023hierarchical} & S2W & \underline{0.617} & \underline{0.617} & \underline{0.617} & \underline{0.618} & \underline{0.863} \\
        MaLP~\cite{asnani2023malp} & S2W &  0.586 &  0.570 & 0.563 & 0.611 & 0.612 \\
        PADL & S2W & \textbf{0.772} & \textbf{0.742} & \textbf{0.739} & \textbf{0.855} & \textbf{0.921} \\
        \bottomrule
        
    \end{tabular}
    
    \label{tab:generalization}
}

\end{table}

It is possible to appreciate from~\cref{tab:generalization} that passive methods~\cite{dang2020detection,guo2023hierarchical} achieve performance comparable to the state of the art only on GMs used at training time~\cite{choi2020stargan} but are unable to generalize across unseen generative models in both detection and localization.
In particular, images manipulated by more advanced architectures are recognized as real images. 

Compared to both passive and proactive methods~\cite{dang2020detection,guo2023hierarchical,asnani2023malp}, \cref{tab:generalization} shows that PADL achieves more robust performance in both detection and localization, while other solutions fall short in localization when the detection performance decreases.

In addition, PADL is able to identify manipulation even when tested on data from a different domain, as can be appreciated from the performance observed when employing the Summer2Winter dataset.

Finally, PADL achieves remarkable detection performance, with near-perfect accuracy, even against the latest generative models like SD and SDXL, despite being trained only on STGAN, outperforming other solutions by a significant margin. 

\begin{table}
\centering
\caption{\tbf{Detection accuracy with reverse engineered perturbation}. The reverse-engineered perturbation is applied to a set of images which is then fed to the detector of the relative method. A high detection accuracy means that the perturbation has been correctly reverse-engineered, i.e., lower values indicate a more robust approach. The experiments have been conducted using an increasing number of protected images, from 4 up to 64. Results have been averaged across ten trials. }

\resizebox{0.6\textwidth}{!}{
    \begin{tabular}{cccccc}
        \toprule
        \multirow{2}{*}{\tbf{K}} & \multicolumn{2}{c}{\textbf{Attack (\% $\downarrow$)}} & \multicolumn{1}{c}{\textbf{Adaptive Attack (\% $\downarrow$) }} \\
        \cmidrule(lr){2-3} \cmidrule(lr){4-4}
          &  MaLP~\cite{asnani2023malp} &   PADL &  PADL \\
        \midrule
         4 & 0.982 $\pm$ 0.020  & 0.001 $\pm$ 0.003  & 0.004 $\pm$ 0.005 \\
         8 & 0.971 $\pm$ 0.016  & 0.019 $\pm$ 0.014  & 0.004 $\pm$ 0.004 \\
        16 & 0.979 $\pm$ 0.012  & 0.002 $\pm$ 0.002  & 0.004 $\pm$ 0.005 \\
        32 & 0.975 $\pm$ 0.011  & 0.028 $\pm$ 0.002  & 0.007 $\pm$ 0.007 \\
        64 & 0.981 $\pm$ 0.011  & 0.012 $\pm$ 0.005  & 0.002 $\pm$ 0.003 \\
        \bottomrule    
        \end{tabular}
}\label{tab:reverse}
\end{table}

\subsection{Black-box attack to proactive scheme: reverse engineering of the protection}
\label{sec:reverse}

To assess the safety related to using the same protection for all the images, we designed a simple attack, performed in a black-box scenario, i.e., without knowledge of the detection model, to extract the perturbation from a limited number of protected images. 
This can be later exploited to deceive the detection model with new, unseen and unprotected images. 
The attack leverages a dataset composed by $K$ protected images $\prot(\bx)$, taken from CelebA and a set of different unprotected images $\bx$, taken from CelebA-HQ. These images have been selected from different datasets to maximize the fairness of this experiment.
The architecture is composed by a learnable perturbation $\pert$ and a CNN model $\mathcal{P}_{ext}$ which serves as protection extractor. 
Given $\prot(\bx) = \bx + \pert_e$, we seek to estimate the unknown perturbation $\pert_e$ by decomposing $\prot(\bx)$.
During training, the learnable perturbation $\pert$ and the CNN model $\mathcal{P}_{ext}$ are jointly optimized using the following loss:
\begin{equation}
    \Loss_{\text{attack}} = \cosine{\pert}{\mathcal{P}_{ext}(\prot(\bx)))} + ||\mathcal{P}_{ext}(\bx)||_2 + ||\pert||_2
  \label{eq:attack_loss}
\end{equation}

The first term of the loss computes the cosine similarity between the learnable $\pert$ and the extracted perturbations, estimated by $\mathcal{P}_{ext}$, thereby constraining the protection to resemble the perturbation structure. 
The second term computes the $\ell_2$ norm of unprotected images and is used to force the protection extractor to focus on the estimation of $\pert$ instead of being guided by the image content. 
Finally, the third term aims at minimizing the magnitude of the reversed perturbation via an $\ell_2$ loss term so as to mitigate potential degradation in the quality of the image.
Once the perturbation $\pert$ has been optimized it can be applied to a new set of unprotected images. 
For this experiment, we employ the test set defined from FFHQ~\cite{karras2019stylebased} and used for the generalization experiment so as to ensure no overlap with the images seen during training.
The reversed perturbation is applied to these images which are then provided as input to the detection model in order to predict if they are protected or not.

The experiment was conducted using an increasing number of protected images (e.g., from a minimum of 4 up to 64). In addition, the results of this experiment were averaged over 10 trials, and for each trial, a different subset of protected images was considered.
From~\cref{tab:reverse}, it is possible to appreciate that the proposed attack successfully estimates the fixed protection proposed by~\cite{asnani2023malp} resulting in 98\% of accuracy. The learned protection $\pert$ is capable of approximating the original perturbation $\pert_e$ with an average cosine similarity of $0.76$ across all $K$.
Conversely, when the attack is applied to our solution, thanks to the image-specific protection, the accuracy drops drastically, meaning that it is not possible to estimate a perturbation capable of breaking our model. 
The same protocol was employed to attack~\cite{asnani2022proactive}. Results showed a constant accuracy close to 100\% across all values of $K$. 
These results can be attributed to the inherent lack of robustness of these models since they were not designed to be robust and accept random noise as a protection, a flaw that can be exploited in the attack as shown in~\cref{fig:reverse_result}(a).

To further stress our approach, we additionally designed a black-box adaptive attack~\cite{tramer2020adaptive}, specifically tailored to our approach. 
Similarly to the previous attack, we employ the same CNN-based model as the protection extractor, however, rather than relying on a singular learnable perturbation, we employ a set of perturbations, proportional to the number of protected images, to better accommodate the inherent diversity of our protection mechanism. Additionally, a perturbation diversity loss, $\Loss_{div}$, is applied to the ensemble of perturbations to enforce variance within the set. As reported in \cref{tab:reverse}, despite the adaptive attack, our model demonstrates its robustness, yielding comparable results to those observed for the previous attack.

\subsection{Protection impact on image quality}
\begin{figure}
    \centering
    \begin{overpic}[width=0.95\textwidth]{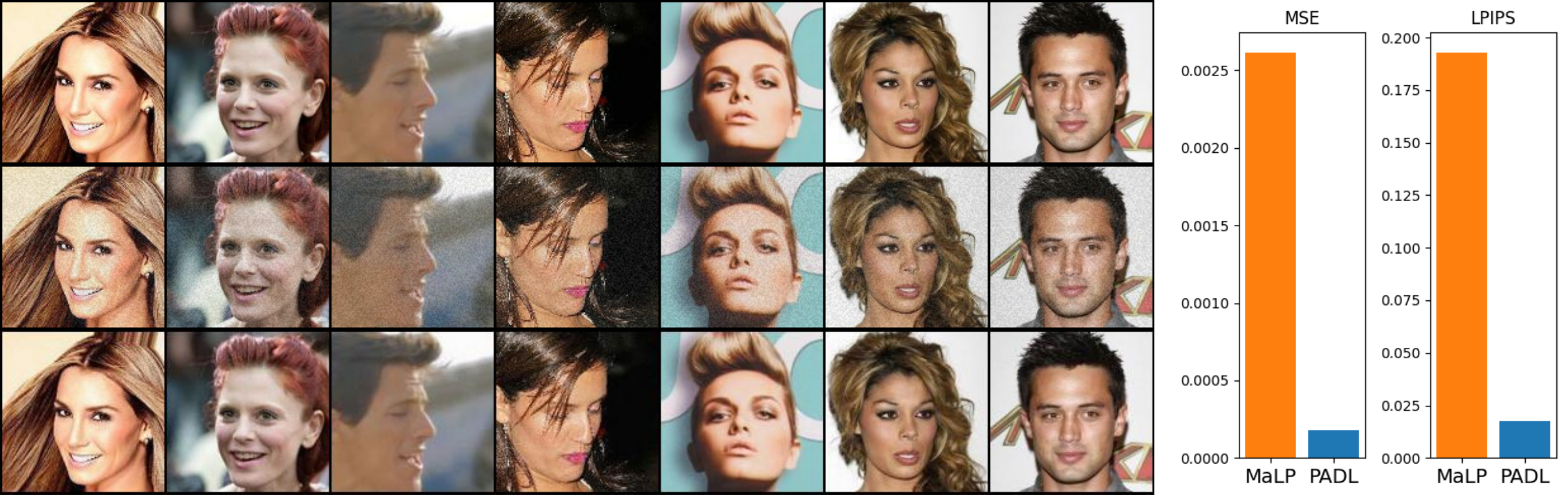}
    \put(-24,230){\rotatebox{90}{\small{{Input}}}}
    \put(-24,25){\rotatebox{90}{\small{{PADL}}}}
    \put(-24,110){\rotatebox{90}{\small{{MaLP~\cite{asnani2023malp}}}}}
    \put(355,-25){\textbf{(a)}}
    \put(870,-25){\textbf{(b)}}
    \end{overpic}
    \vspace{2mm}

    \caption{\tbf{Comparison of protected images.} (a) original images (\emph{top}), images protected by~\cite{asnani2023malp} (\emph{middle}) and by PADL (\emph{bottom}). Zoom in for better visualization. (b) Image quality measured in terms of MSE and LPIPS
both calculated between real images and their protected version.}
    \label{fig:image_comparison}
\end{figure}
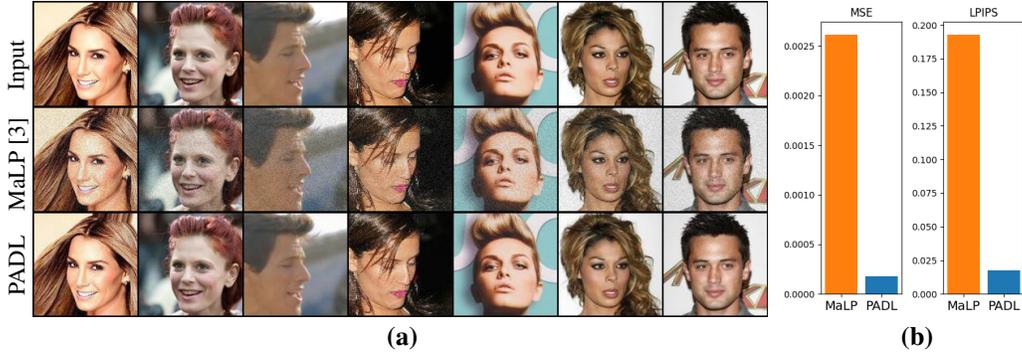

The process of image protection may reduce the quality of the visual output. 
To quantify this phenomenon, we measure the degradation between $\bx$ and $\prot(\bx)$ at the pixel level by computing the mean squared error (MSE), and at the perceptual level, employing the Learned Perceptual Image Patch Similarity~\citep{zhang2018perceptual} (LPIPS).
The results reported in \cref{fig:image_comparison}~(b) confirm that our protection mechanism has little impact on the overall image quality, as highlighted by the very low values for both MSE and LPIPS metrics. 
This result is also supported by the qualitative examples reported in \cref{fig:image_comparison}~(a). Here, protected images are compared to their original input counterparts. The protection applied by~\cite{asnani2023malp} is more noticeable, as can be observed from \cref{fig:image_comparison}~(a) and also by the higher MSE and LPIPS values in \cref{fig:image_comparison}~(b). 
Compared to~\cite{asnani2023malp}, PADL allows better performance, while also minimizing the impact on image quality.
This is a consequence of the fact that, differently from~\cite{asnani2023malp},  our solution applies an upper bound to limit the perturbation by combining the use of hyperbolic tangent and $\alpha$, as described in \cref{sec:encoding_module}. 
Ablation on the impact of the protection strength $\alpha$ on the image quality is also provided in the supplemental material. 

\section{Conclusion}
\label{sec:conclusion}

This work introduces a novel solution for proactive image manipulation detection and localization. 
Our solution employs a transformer-based encoder conditioned by an input image to generate a specific perturbation. Then a transformer-based decoder is used to extract the perturbation and leverage it to perform manipulation localization and detection. 
Unlike previous methods based on fixed protection, our solution generates image-specific perturbations, improving resistance against reversal attacks, while also achieving remarkable detection and localization performance. It is also worth highlighting that the perturbation introduced by our approach has very little impact on the image quality.

\minisection{Broader impact}
The objective of this research is to prevent the misuse of generative image models therefore mitigating the spread of misinformation. 
By enabling a more effective detection of manipulated images, we hope to offer a way to bolster trust and integrity in digital content, which is crucial for fields such as journalism, forensics, and law enforcement. 

\minisection{Limitations and future works}
The main limitation of our solution is related to the drop in performance for the localization when generative models based on a diverse architecture and paradigm (e.g., diffusion model) are employed. 
In order to enhance performance in this regard, it would be interesting to explore the potential of novel architectural approaches for both decoder and encoder modules.
Although our method demonstrates superior performance compared to previous approaches, further investigation is required to assess its suitability for real-world scenarios. Online platforms may apply filters to uploaded images, potentially compromising the embedded protection.
Additionally, it would be worthwhile to assess whether the methodology employed for perturbation generation can be repurposed for other tasks, such as adversarial attacks. These represent promising directions for further research and advancement in the field.

\section*{Acknowledgement}
\label{sec:Acknowledgement}
Funded by the European Union – Next Generation EU within the framework of the National Recovery and Resilience Plan NRRP – Mission 4 "Education and Research" – Component 2 - Investment 1.1 “National Research Program and Projects of Significant National Interest Fund (PRIN)” (Call D.D. MUR n. 104/2022) – PRIN2022 – Project reference: Adversarial Venture, the Mixed Blessing of Adversarial Attacks 20227YET9B\_002 J53D23007030006.

{
\small
\bibliographystyle{splncs04}
\bibliography{egbib}
}

\newpage
\appendix
\section{Supplemental material}

\subsection{Loss Ablation}
\label{sec:ablation}
Compared to previous art~\cite{asnani2022proactive,asnani2023malp}, which utilized up to ten loss functions to achieve their objectives, PADL simplifies the approach by employing only four losses, as detailed in \cref{sec:training}, each specifically designed to enforce essential properties, resulting in a more efficient yet effective model. The removal of any one of these four losses would prevent the model from functioning as intended. For instance, removing the $\mathcal{L}_{div}$ loss would prevent the model from generating image-specific perturbations. Without $\mathcal{L}_{div}$, the model would minimize the remaining losses by learning a single perturbation for all images, which contradicts our design goals.

\begin{figure}[!h]
    \centering
    \includegraphics[width=1\linewidth]{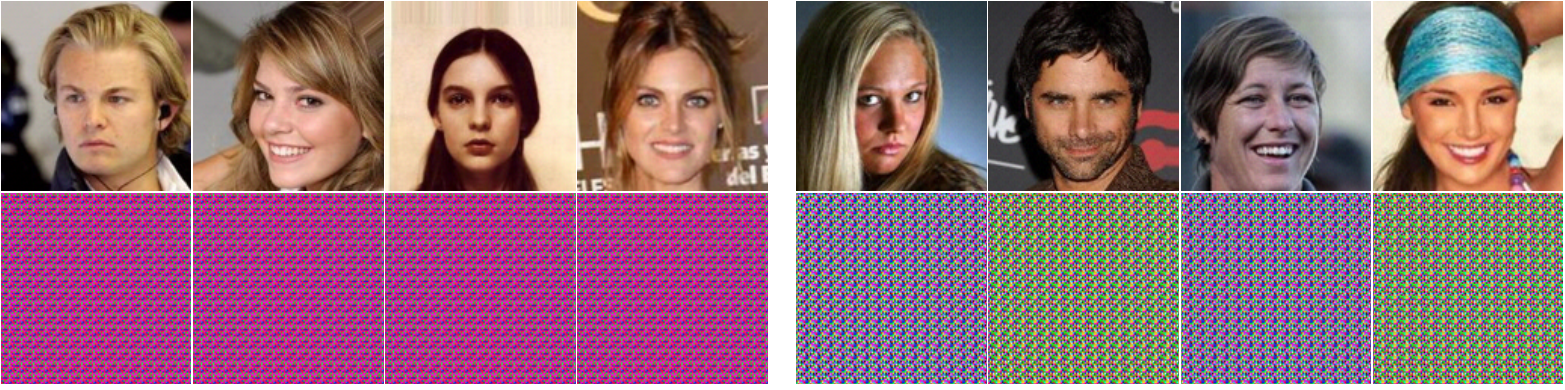}
    \caption{\tbf{Loss Ablation:} (Left) The model is unable to generate different perturbations without $\mathcal{L}_{div}$. (Right) Without the max in $\mathcal{L}_{div}$ the model learns only two perturbations in opposite directions.}
    \label{fig:ablation}
\end{figure}

Moreover, using plain cosine similarity (i.e., $\mathcal{L}_{div}$ without the clipping $\max(\cdot,0)$) failed to produce image-specific perturbations. The encoder ended up learning only two distinct perturbations with a cosine similarity of -1, essentially opposite directions that merely minimized the loss. This led to a situation where, within a batch, the mean cosine similarity approached zero due to the compensatory effect between similar and opposite perturbations, resulting in no meaningful learning. To counter this, negative values were clamped to zero, effectively disregarding pairs with negative similarity and forcing the perturbations to be orthogonal. Additional evidence of this behavior is provided in \cref{fig:ablation}.

\subsection{Training PADL with a Diffusion Model}
\label{sec:padl-diff}

To provide additional evidence of the capability of our solution, we trained an additional version of PADL using a diffusion model as the GM. Results in \cref{tab:padl-diff} show that the model continues to perform robustly against unseen manipulations (SDXL, BlendGAN, and StarGANv2), demonstrating its strong generalization capability.

\begin{table}[h]
    \centering
    \caption{\tbf{Performance of PADL trained with Stable Diffusion 1.5 as the Generative Model.} Experiments are conducted with the PADL model using $N = 3$ layers. ``t'' represents the threshold used to binarize the GT masks. The dataset used is CelebA-HQ.}
    \resizebox{0.8\textwidth}{!}{
    \begin{tabular}{cccccc}
        \toprule
        \multirow{2}{*}{\textbf{Model}} & \multicolumn{3}{c}{\textbf{Localization}} & \multicolumn{2}{c}{\textbf{Detection}} \\
        \cmidrule(lr){2-4} \cmidrule(lr){5-6}
          & AUC t=0.1 ($\uparrow$) & AUC t=0.25 ($\uparrow$) & AUC t=0.5 ($\uparrow$) & Acc. ($\uparrow$) & AUC ($\uparrow$) \\

        \midrule
         
        & &  \multicolumn{4}{c}{\tbf{StarGANv2}} 
        \\
        \cmidrule(lr){1-1}
        \cmidrule(lr){2-4}
        \cmidrule(lr){5-6}

        PADL Diff. & 0.754 & 0.698 & 0.692 &  0.820 & 0.985\\
        PADL STGAN & \textbf{0.933} & \textbf{0.868} & \textbf{0.835} & \textbf{0.985} & \textbf{1.00} \\
        \midrule

        &&\multicolumn{4}{c}{\tbf{BlendGAN}} \\
        \cmidrule(lr){1-1}
        \cmidrule(lr){2-4}
        \cmidrule(lr){5-6}
        
        PADL Diff.  & 0.928 & 0.843 & 0.790 &\textbf{1.00} & \textbf{1.00} \\
        PADL STGAN  & \textbf{0.941} & \textbf{0.871} & \textbf{0.798} & \textbf{0.997} & \textbf{1.00}\\
        \midrule

        &&\multicolumn{4}{c}{\tbf{SD 1.5}} \\
        \cmidrule(lr){1-1}
        \cmidrule(lr){2-4}
        \cmidrule(lr){5-6}

        PADL Diff.  & \textbf{0.897} & \textbf{0.909} & \textbf{0.941 }& \textbf{1.00} &\textbf{1.00} \\
        PADL STGAN & 0.794 & 0.766 & 0.775 & 0.997 & \textbf{1.00}\\
        \midrule

        &&\multicolumn{4}{c}{\tbf{SDXL}} \\
        \cmidrule(lr){1-1}
        \cmidrule(lr){2-4}
        \cmidrule(lr){5-6}

        PADL Diff.  & \textbf{0.910 }   & \textbf{0.899}  & \textbf{0.917} &  \textbf{1.00} & \textbf{1.00} \\
        PADL STGAN  & 0.812   & 0.773 & 0.775 & 0.997 & \textbf{0.999} \\

        \bottomrule
    \end{tabular}
    \label{tab:padl-diff}
}
\end{table}

Testing PADL trained with diffusion against DiffAE and STGAN was not possible due to resolution incompatibility. PADL maintains strong generalization performance on unseen GMs, proving that its effectiveness lies in the method itself, not the specific training data used.

\subsection{Robustness against degradations}
Both passive and proactive methods may fall short when the images undergo some simple editing degradations. 
To this end, we conducted an experiment to evaluate the performance of our approach when four types of degradations are applied, namely blur, Gaussian noise, JPEG compression, and low resolution. This experiment has been conducted following a leave-one-out protocol, that is, we trained four model injecting three out of the four degradations during training and we tested on the unseen degradation.
More in details, during training we adopted a degradation scheduler policy that challenges the optimization of the perturbation.  At each iteration, it randomly selects whether training will proceed with original images or with one of the three degradations. The probability of employing a degradation and its intensity follows a linear schedule, which is proportional to the current iteration count.

It is worth noting that these degradations are applied directly on protected images to resemble a real-world scenario.
Moreover, the images have been manipulated using STGAN~\citep{liu2019stgan}, in particular, by modifying the ``Bald'' attributes. 

Following the work of~\citep{asnani2023malp, fakelocator} we employed the following settings for the degradations:
\begin{enumerate}
    \item \textbf{Compression}: Images are saved as JPEG with a quality level set to 50\%.
    \item \textbf{Blur}: A Gaussian blur is applied to the images using a $7 \times 7$ kernel.
    \item \textbf{Noise}: Gaussian noise with zero mean and unit variance is added to the images. To preserve the unity gain, values over $1$ and below $-1 $ are clamped.
    \item \textbf{Low-Resolution}: The image is resized to half of its original resolution and then upscaled back to the original resolution using bilinear upsampling.
\end{enumerate}

As reported in \cref{tab:degradations}, our solution is susceptible to two degradations, such as, JPEG compression and Gaussian noise. This is due to the fact that both these degradations significantly compromise the quality of the protected image, leading to its detection as manipulated.
To enhance the overall robustness of the framework we also trained PADL considering all degradations during training.

\begin{table}[H]
    \centering
    \caption{PADL performance against diverse image degradations.}
    \small
    \resizebox{0.8\textwidth}{!}{
    \begin{tabular}{cccccc}
        \toprule
        \multirow{2}{*}{\textbf{Degradations}} &  \multicolumn{3}{c}{\textbf{Localization}} & \multicolumn{2}{c}{\textbf{Detection}} \\
        \cmidrule(lr){2-4} \cmidrule(lr){5-6}
        & AUC t=0.1 ($\uparrow$) & AUC t=0.25 ($\uparrow$) & AUC t=0.5 ($\uparrow$) & Acc. ($\uparrow$) & AUC ($\uparrow$) \\

        \midrule
        \multicolumn{6}{c}{Leave-one-out experiment}\\
        \midrule
        Compression & 0.250 & 0.404 & 0.577 & 0.502 & 0.833 \\
        \cmidrule(lr){1-1}
        \cmidrule(lr){2-4}
        \cmidrule(lr){5-6}
        Blur  & 0.744 & 0.853 & 0.957 & 0.748 & 0.999 \\
        \cmidrule(lr){1-1}
        \cmidrule(lr){2-4}
        \cmidrule(lr){5-6}
        Noise  & 0.113 & 0.291 & 0.488 & 0.485 & 0.967 \\
        \cmidrule(lr){1-1}
        \cmidrule(lr){2-4}
        \cmidrule(lr){5-6}
        Low Res. & 0.865 & 0.932 & 0.985 & 0.873 & 1.00 \\

        \midrule
        \multicolumn{6}{c}{Training with all degradations}\\
        \midrule

        Compression & 0.857 & 0.85 & 0.913 & 0.953 & 0.991 \\
        \cmidrule(lr){1-1}
        \cmidrule(lr){2-4}
        \cmidrule(lr){5-6}
        Blur & 0.732 & 0.862 & 0.967 & 0.720 & 0.998 \\
        \cmidrule(lr){1-1}
        \cmidrule(lr){2-4}
        \cmidrule(lr){5-6}
        Noise & 0.947 & 0.858 & 0.885 & 1.00 & 1.00 \\
        \cmidrule(lr){1-1}
        \cmidrule(lr){2-4}
        \cmidrule(lr){5-6}
        Low Res. & 0.751 & 0.872 & 0.971 & 0.743 & 0.996 \\

        \bottomrule
    \end{tabular}
    }
    \label{tab:degradations}
\end{table}

All models reported in \cref{tab:generalization,tab:padl_performance,tab:reverse,tab:padl-diff} have been trained considering all degradations.

\subsection{Perturbation strength}
\begin{figure}[h]
    \centering
    \begin{overpic}[width=1\textwidth]{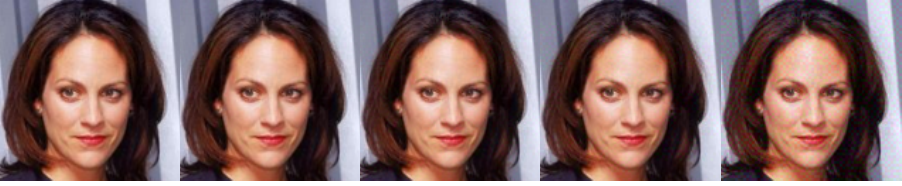}
        \put(22,-15){Original Image}
        \put(255,-15){$\alpha$ = 0.01}
        \put(450,-15){$\alpha$ = 0.03}
        \put(655,-15){$\alpha$ = 0.05}
        \put(855,-15){$\alpha$ = 0.1}
        
    \end{overpic}
    \vspace{0.5mm}
    \caption{\tbf{Qualitative comparison of protected images with different protection strengths.} Progression of the visual quality of protected images with an increasing value of $\alpha$. Values over $0.05$ result in visible artifacts which compromise the image quality. Zoom in for better visualization.}
    \label{fig:alpha_comparison}
\end{figure}

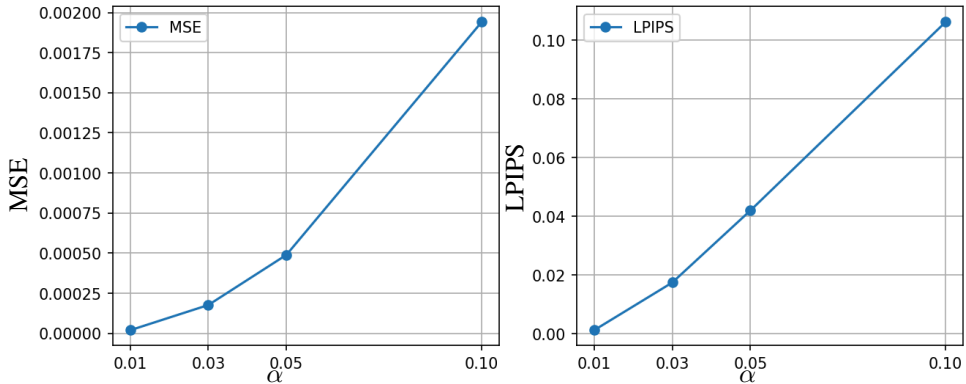
\begin{figure}[h]
    \centering
    \begin{overpic}[width=0.9\textwidth]{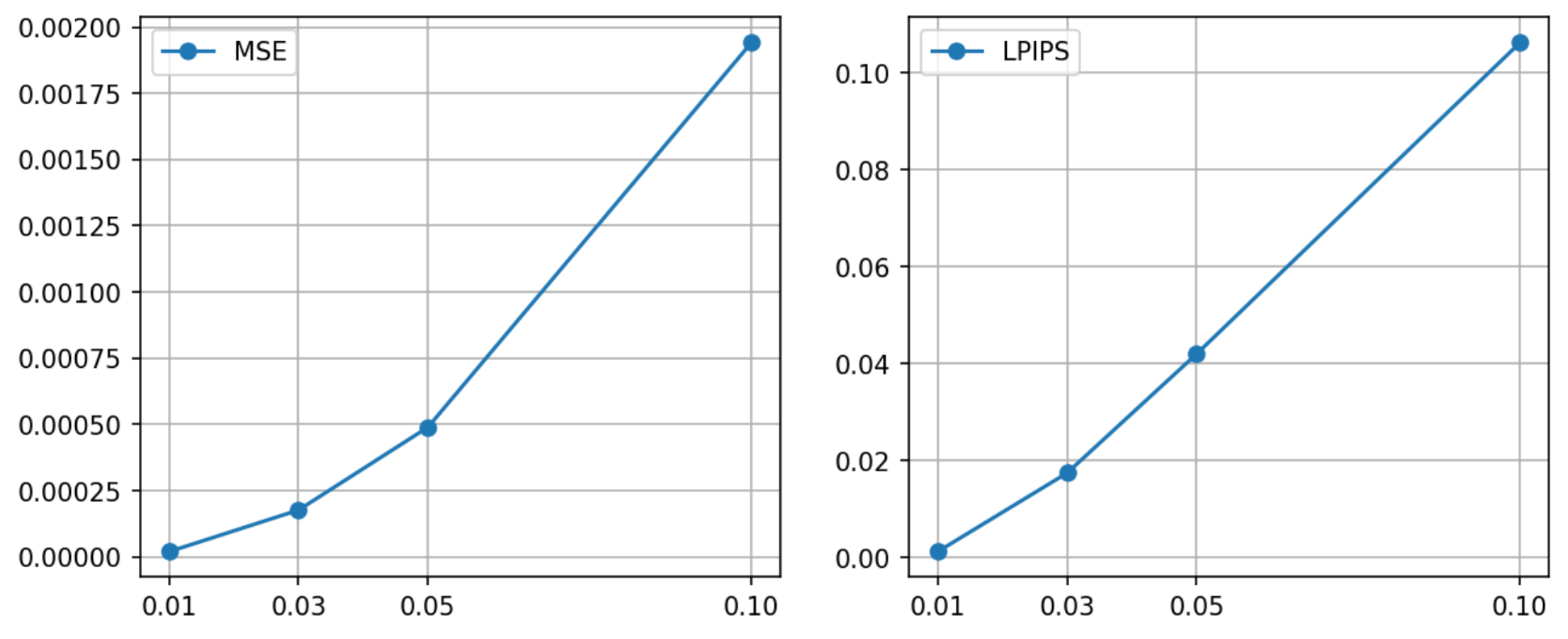}
        \put(252,0){$\alpha$}
        \put(752,0){$\alpha$}
        \put(-20,180){\rotatebox{90}{{{MSE}}}}
        \put(505,180){\rotatebox{90}{{{LPIPS}}}}
    \end{overpic}
    \caption{\t{Quantitative comparison of protected images with different protection strengths.} MSE and LPIPS results for an increasing value of $\alpha$.}
    \label{fig:alpha_quantitative_comparison}
\end{figure}

Variations in the $\alpha$ parameter directly correspond to proportional changes in the magnitude of the perturbation. 
Employing a hyperbolic tangent on the output of $\pertenc$ allows us to bound the maximum value of the perturbation, consequently, $\alpha$ is the only parameter that controls the magnitude. 
To show the impact of alpha on the image quality we conducted an experiment by training four models with different values of $\alpha$. 
Quantitative results are reported in~\cref{fig:alpha_quantitative_comparison} while some qualitative samples are shown in \cref{fig:alpha_comparison}. 
Using a value for $\alpha$ higher than $0.03$ results in visible artifacts that degrade the image quality.
For all our experiments we set $\alpha=0.03$ since it represents the optimal balance between quality (i.e., the perturbation magnitude is sufficiently low to preserve image quality) and performance (i.e., the magnitude is sufficiently high to ensure detectability by the decoder).

\subsection{Perturbation variation across model depths}
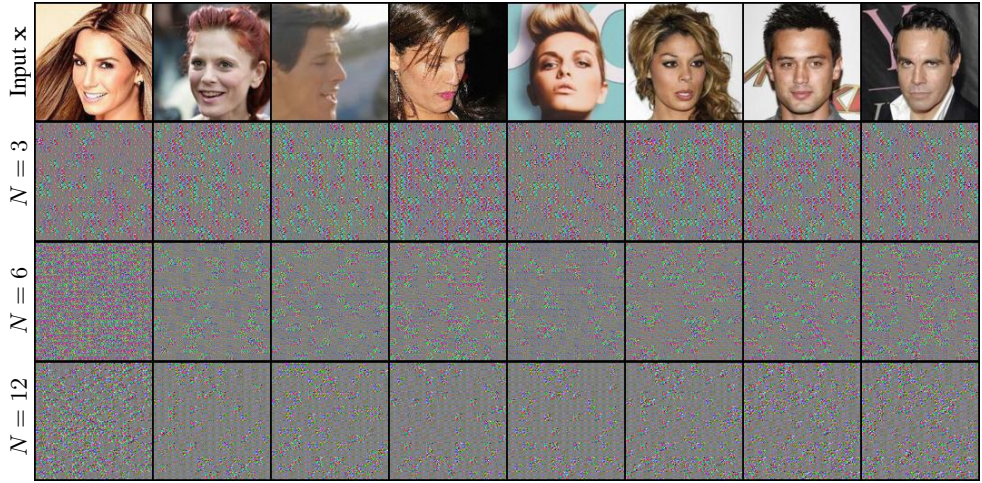
\begin{figure*}
    \centering
    \begin{overpic}[width=0.9\textwidth]{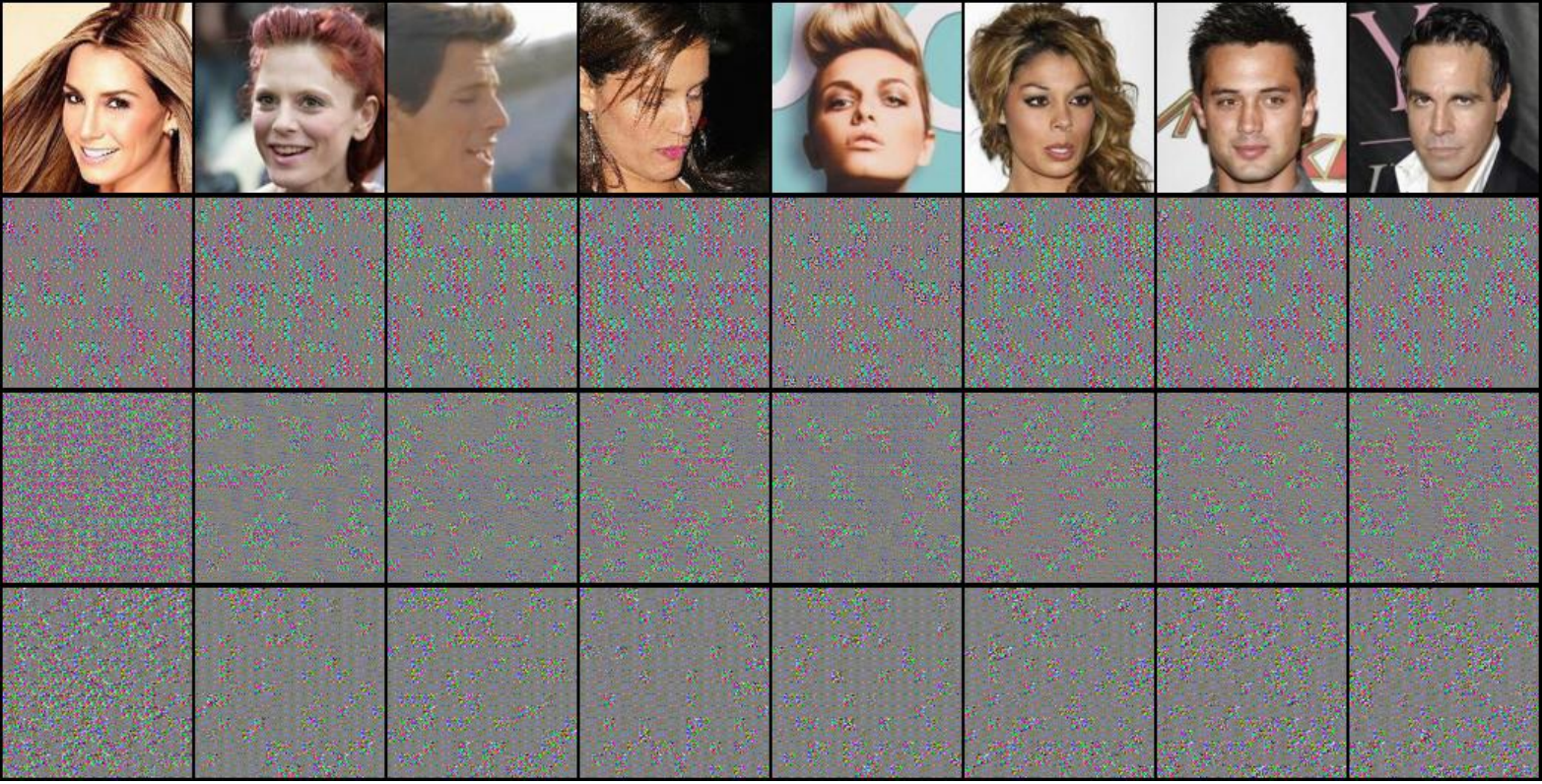}
        \put(-25,408){\rotatebox{90}{\small{Input  $\bx$}}}
        \put(-25,292){\rotatebox{90}{\small{$N = 3$}}}
        \put(-25,158){\rotatebox{90}{\small{$N = 6$}}}
        \put(-25,28) {\rotatebox{90}{\small{$N = 12$}}}    
    \end{overpic}
    \caption{Visual representation of the perturbations for different images and considering models with an increasing number of transformer layers. For visualisation purposes, all perturbations are normalised between $[0,1]$.}
    \label{fig:pert_depth}
\end{figure*}

A different number of transformer layers can influence the generated perturbation. 
This phenomenon is shown in~\cref{fig:pert_depth}.
The perturbations across models with a different number of layers are visually similar, yet different across images. 
It is possible to notice that a patch-based pattern emerges mainly because of the way the transformer architecture processes the images. 
However, the pattern of the patches becomes more complex as the depth increases. 
Although the $N=12$ model produces a markedly distinctive appearance for each patch, it is evident that even the shallower model ($N=3$) can generate perturbations which are different across images.

\subsection{Reverse attack with multiples templates}
\begin{figure}[h]
    \centering
    \begin{overpic}[width=0.90\textwidth]{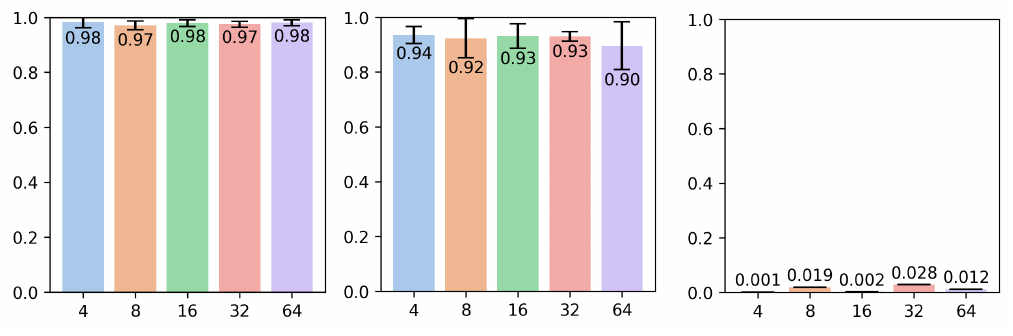}
        \put(-25,130){\rotatebox{90}{{{Accuracy}}}}
        
        \put(190,-5){\small{K}}
        \put(507,-5){\small{K}}
        \put(841,-5){\small{K}}
        
        \put(155,-32){\small{MaLP~\cite{asnani2023malp}}}
        \put(380,-32){\small{MaLP~\cite{asnani2023malp} w/ 3 perturbations}}
        \put(820,-32){\small{PADL}}
    \end{overpic}
    \vspace{2mm}
    \caption{\tbf{Detection accuracy with reverse-engineered perturbation}. The reverse-engineered perturbation is applied to a set of images which is then fed to the detector of the relative method. A high detection accuracy means that the perturbation has been correctly reverse-engineered, i.e., lower values indicate a more robust approach. The experiments have been conducted using an increasing number of protected images, from 4 up to 64. Results have been averaged across ten trials.}
    \label{fig:reverse_3t}
\end{figure}
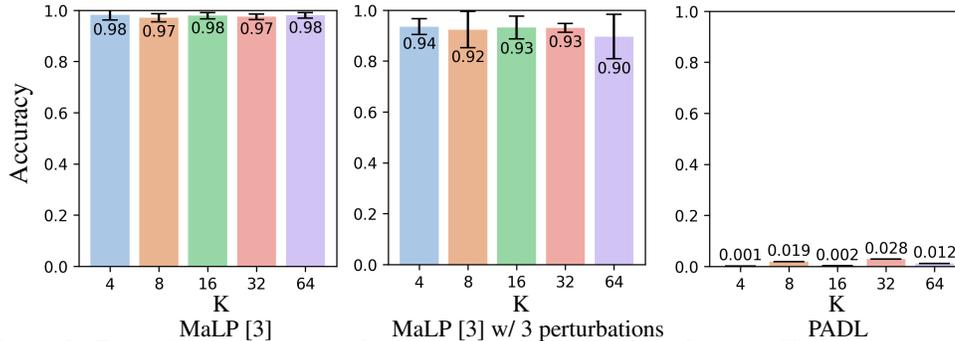

The solution proposed by~\citep{asnani2023malp} (MaLP) released only the model with a single perturbation, the attack described in~\cref{sec:reverse} has been conducted using this model. 
In order to better ascertain the proposed attack, we conducted an additional test, training their model using three perturbations, since it was shown to achieve the highest performance in~\citep{asnani2023malp}. 
As is possible to observe from~\cref{fig:reverse_3t}, even with a set of three perturbations, MaLP~\cite{asnani2023malp} remains susceptible to reverse attacks.

\subsection{Visualization of manipulation maps}
\begin{figure}
    \centering
    \begin{overpic}[width=0.9\textwidth]{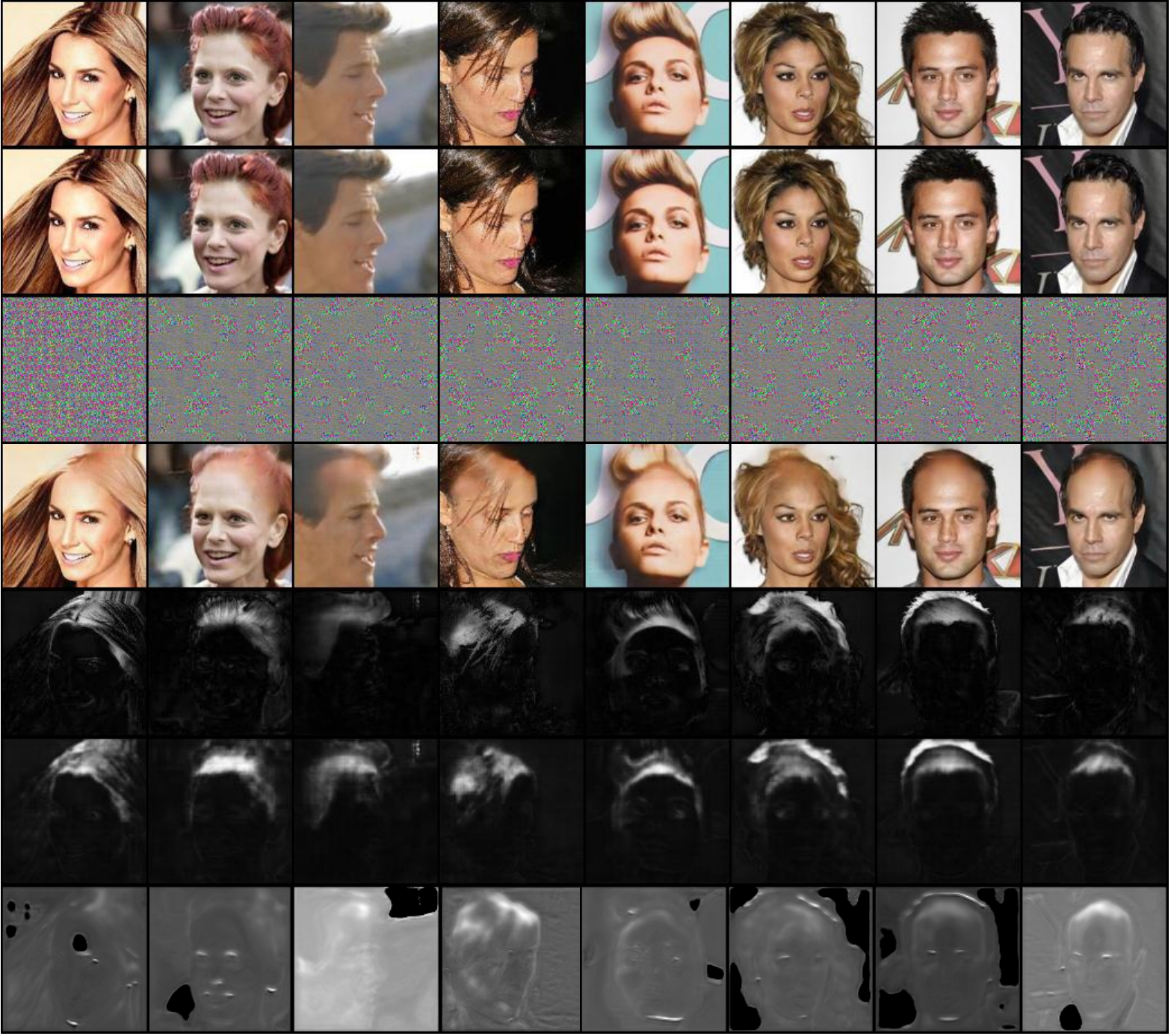}
        \put(-25,815){\rotatebox{90}{\small{{$\bx$}}}}
        \put(-30,675){\rotatebox{90}{\small{{$\prot(\bx)$}}}}
        \put(-30,545){\rotatebox{90}{\small{{$\pert(\bx)$}}}}
        \put(-30,420){\rotatebox{90}{\small{{$\gen(\bx)$}}}}
        \put(-30,275){\rotatebox{90}{\small{{Map GT}}}}
        \put(-30,175) {\rotatebox{90}{\small{{Map}}}}    
        \put(-30,35) {\rotatebox{90}{\small{{MaLP}}}}    
    \end{overpic}
    \caption{Visualization of manipulation maps of STGAN}
    \label{fig:stgan_examples}
\end{figure}

\cref{fig:stgan_examples} illustrates a selection of real images, accompanied by their protected version, the image-specific perturbations and the manipulated versions, along with the ground truth and the estimated manipulation maps of PADL and MaLP~\cite{asnani2023malp}.
STGAN manipulations are local to specific attributes of the image and this clearly influences the look of the map estimated by our model.

\subsection{Visualization of PADL predictions}
\begin{figure}[h]
    \centering
    \begin{overpic}[width=0.90\textwidth]{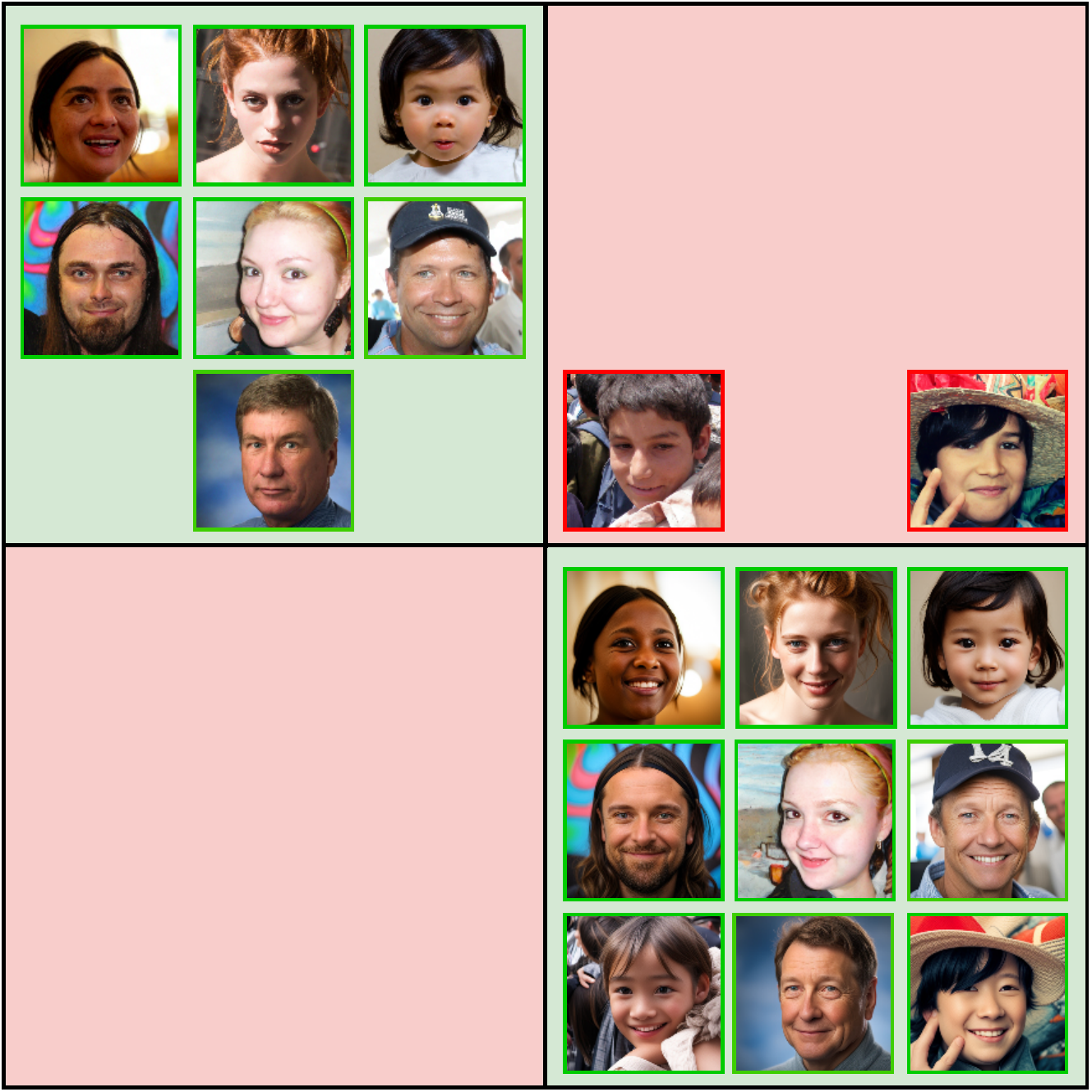}

    \put(410, 1030){\large{\textbf{Ground Truth}}}
    \put(185, 1010){\large{\textit{Protected}}}
    \put(685, 1010){\large{\textit{Manipulated}}}

    \put(-55, 410){\rotatebox{90}{\large{\textbf{Prediction}}}}
    \put(-30,185){\rotatebox{90}{\large{\textit{Manipulated}}}}
    \put(-30,685){\rotatebox{90}{\large{\textit{Protected}}}}
    
    \put(95, 265){\large{No protected image was}}
    \put(95, 230){\large{detected as manipulated.}}
    
    \end{overpic}
    \caption{Visualization of correct and wrong predictions: Images displayed in green are correctly detected, while red indicates incorrectly detected images.}
    \label{fig:sdxl_examples}
\end{figure}
In~\cref{fig:sdxl_examples} we present a selection of images generated using SDXL, the most recent and advanced generative model among those employed in our evaluation, which is able to generate images that look real to the human eye.
Nonetheless, PADL, which is trained only on older GAN-based generative models, is able to correctly identify extremely realistic manipulations with accuracies approaching $100\%$.

\subsection{Additional Implementation details}
The STGAN model has been detached from the computational graph, therefore its gradient is not exploited during the training process. 
Consequently, the models are unable to rely on the STGAN architecture during training, which results in a solution capable of generalizing across both detection and localization.

The evaluation of the state-of-the-art models was conducted by utilising the original code and models released by the authors. The results of~\citep{asnani2023malp}'s cosine similarity values do not correspond with those presented in their original article. This discrepancy is caused by a calculation error in the ground truth of the manipulation maps, which was present in the released code. To ensure fairness, all their values have been recalculated.

\subsection{Generative Models, datasets and relative licenses}
For each GM used at test time we employed the reference test set released by~\citep{asnani2023malp}. 
Each test set corresponds to a subset of 200 real image taken from the original source dataset. 
As new GMs were introduced, we complemented the test set images with new images from the FFHQ dataset~\citep{karras2019stylebased}.
For CelebA~\citep{liu2015deep} and CelebA~HQ~\citep{karras2018progressive}, we use the test images released by~\citep{asnani2022proactive,asnani2023malp}. 
For a fair comparison, we will release our new test images based on FFHQ~\citep{karras2019stylebased}.

\begin{table}[h]
    \centering
    \caption{List of used GMs.}
    \resizebox{\textwidth}{!}{
    \begin{tabular}{cccc}
        \toprule
        {\textbf{Model}} & 
        {\textbf{Architecture}} &
        {\textbf{Task}} & 
        {\textbf{License}}\\
        \midrule
        STGAN~\citep{liu2019stgan} & GAN~\citep{goodfellow2014generative}  & Attribute Manipulation  & MIT\\
        StarGANv2~\citep{choi2020stargan} & GAN~\citep{goodfellow2014generative}  & Style transfer & CC-BY-NC 4.0\\
        BlendGan~\citep{liu2021blendgan} & GAN~\citep{goodfellow2014generative}  & High resolution style transfer &  MIT\\
        DiffAE~\citep{diffae} & DDIM~\citep{song2022denoising} & Attribute Manipulation & MIT\\
        StableDiffusion 1.5~\citep{SD15} & Latent Diffusion~\citep{rombach2022highresolution} & Img2Img & CreativeML Open RAIL-M \\
        StableDiffusion XL ~\citep{sdxl} & Latent Diffusion~\citep{rombach2022highresolution} & Img2Img & CreativeML Open Rail++-M \\
        
        \bottomrule
    \end{tabular}
    }
\label{tab:models&datasets}
\end{table}

In the context of image manipulation, Style Transfer models (BlendGAN and StarGANv2) were employed to generate images based on a fixed reference style image. In contrast, the attribute manipulation model DiffAE was configured to alter the facial attribute \emph{``bald''}. The SD and SDXL models were utilized in an img2img configuration, conditioned with the prompt \emph{``a nice picture of a smiling person''} for CelebA-HQ and FFHQ and "a nice picture of a winter landscape with snowy weather" for Summer2Winter.

\cref{tab:models&datasets} provides a list of all generative models used in our experiments, along with their architecture, task and license.
CelebA~\citep{liu2015deep} is intended for non-commercial research purposes only, to which we strictly adhere. Similarly, CelebA HQ~\citep{karras2018progressive} and FFHQ~\citep{karras2019stylebased} are licensed under CC BY-NC-SA 4.0, indicating their availability for non-commercial purposes.

\end{document}